\documentclass[journal]{IEEEtran}
\usepackage{times}
\usepackage{epsfig}
\usepackage{graphicx}
\usepackage{amsthm,amsmath,amssymb}
\usepackage{mathrsfs}
\usepackage{multirow}
\usepackage{makecell}
\usepackage{enumerate}
\usepackage{algorithm}
\usepackage{algorithmic}
\usepackage{color}
\usepackage{stfloats}
\usepackage{bm}
\usepackage{comment}
\usepackage{setspace}

\ifCLASSINFOpdf
\else
\fi
\begin{document}
%
\title{End-to-End Context-Aided Unicity Matching for Person Re-identification}
%
%
%

\author{Min~Cao,
        Cong~Ding,
        Chen~Chen,
        Junchi~Yan,~\IEEEmembership{Member,~IEEE,}
        and~Qi~Tian
\thanks{Chen Chen is the corresponding author (email: chen.chen@ia.ac.cn).}
\thanks{
Min Cao and Cong Ding are with School of Computer Science and Technology, Soochow University, Suzhou 215006, China (email: mcao@suda.edu.cn).}
 \thanks{
Chen Chen is with the Institute of Automation
Chinese Academy of Sciences, Beijing 100190, China (email: chen.chen@ia.ac.cn).}

\thanks{
Junchi Yan is with the Department of Computer Science and Engineering, and MoE
Key Lab of Artificial Intelligence, AI Institute, Shanghai Jiao Tong University,
Shanghai, 200240, China (email: yanjunchi@sjtu.edu.cn).
}
\thanks{Qi Tian is with the Huawei Noah’s Ark Lab, Shenzhen, 518129, China (emial:tianqi1@huawei.com ).}}


%
%

\markboth{IEEE/CAA JOURNAL OF AUTOMATICA SINICA,~Vol.~X, No.~X, X~X}%
{Shell \MakeLowercase{\textit{et al.}}: Bare Demo of IEEEtran.cls
for Journals}
%



\maketitle

\begin{abstract}
Most existing person re-identification methods compute the matching relations between person images across camera views based on the ranking of the pairwise similarities. This matching strategy with the lack of the global viewpoint and the context’s consideration inevitably leads to ambiguous matching results and sub-optimal performance. Based on a natural assumption that images belonging to the same person identity should not match with images belonging to multiple different person identities across views, called the unicity of person matching on the identity level, we propose an end-to-end person unicity matching architecture for learning and refining the person matching relations. First, we adopt the image samples’ contextual information in feature space to generate the initial soft matching results by using graph neural networks. Secondly, we utilize the samples’ global context relationship to refine the soft matching results and reach the matching unicity through bipartite graph matching. Given full consideration to real-world person re-identification applications, we achieve the unicity matching in both one-shot and multi-shot settings of person re-identification and further develop a fast version of the unicity matching without losing the performance. The proposed method is evaluated on five public benchmarks, including four multi-shot datasets MSMT17, DukeMTMC, Market1501, CUHK03 and a one-shot dataset VIPeR. Experimental results show the superiority of the proposed method on performance and efficiency.
\end{abstract}

\begin{IEEEkeywords}
Person re-identification, unicity matching, end-to-end architecture, contextual information, graph neural network, bipartite graph matching.
\end{IEEEkeywords}

%
\IEEEpeerreviewmaketitle

\section{Introduction}

\IEEEPARstart{P}{erson} re-identification (ReID)~\cite{ye2021deep} plays an important role in the intelligent video monitoring system. Given a probe image, person ReID aims to identify the image of the same person from a set of gallery images across camera views. Due to the changes in pose, illumination and background across views, person ReID is a challenging task.

\begin{figure}[tb!]
\begin{center}
   \includegraphics[width=0.9\linewidth, height=0.63\linewidth]{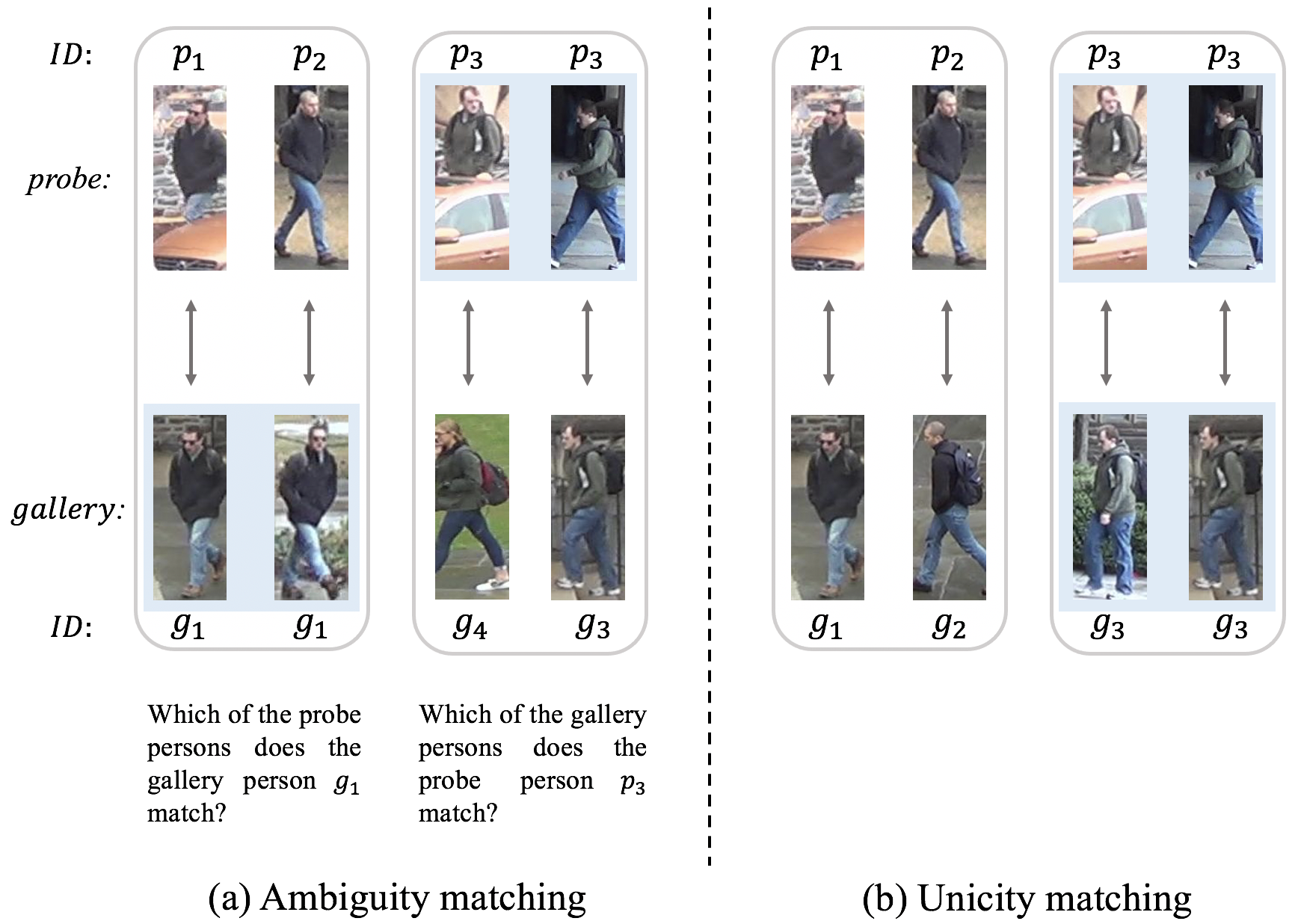}
\end{center}
   \caption{The illustration of the ambiguity matching and unicity matching results on the identity level for person ReID. We show the examples in the DukeMTMC~\cite{ristani2016performance}. In each example, the probe image above and the gallery image below are linked with an arrow as a probe-gallery matching, and the images with the same background color share the same person identity. We also show the identity of the person in the image. The matching results in (a) are obtained by selecting the $1$-th gallery image in the ranking list from the BoT method~\cite{Luo_2019_CVPR_Workshops}. The matching results in (b) are the ground-truth results.}
\label{fig1}
\end{figure}

Essentially, person ReID can be regarded as a person matching task, concentrating on achieving as many correct probe-gallery matchings as possible.
The previous person ReID methods~\cite{matsukawa2016hierarchical,RN2020,ye2021deep} focus on extracting discriminative features or learning proper similarity metrics. 
The probe image's ranking list is computed by sorting the similarity values between the probe image and the gallery images in descending order, 
and we obtain the probe-gallery matching by linking this probe image with the $k$-th gallery image in the list.
However, the matching results are prone to be characterized by ambiguity.
An illustration is shown in Fig.~\ref{fig1} (a). 
The matched gallery images for these probe images are obtained by the Rank-1 results from the state-of-the-art person ReID method~\cite{Luo_2019_CVPR_Workshops}.
The results in Fig.~\ref{fig1} (a) indicate that the same gallery (resp. probe) person matches with two different probe (resp. gallery) persons, which are obviously paradoxical and ambiguous matching results.
By contrast, as shown in Fig.~\ref{fig1} (b), the ideal matching results should be characterized by the unicity of matching, i.e., one probe (resp. gallery) person only matches with one gallery (resp. probe) person who shares the same identity.
The reason for the matching ambiguity is because the matching results are computed solely based on the similarity between the two images, and the similarity relationship in context or global perspective is ignored. 
It can be regarded as a `selfish' matching strategy, being more likely to result in inferior performance. 
To relieve the issue, we should consider the contextual information when computing the matching result in person ReID. 

Several methods~\cite{cao2017from,luo2019spectral,RN2020,shen2018person} have been proposed by utilizing contextual information to achieve the Re-ID performance enhancement. 
However, most of them either aim at learning discriminative features~\cite{si2018dual,zhou2018graph,shen2018person}, or focusing on refining the pairwise similarity by the re-ranking technology~\cite{garcia2017discriminant,cao2020} with the aid of contextual information. The probe-gallery matching results are then obtained still by the `selfish' matching strategy and are characterized by ambiguity.
To the best of our knowledge so far, there are only two methods~\cite{das2014consistent,rezatofighi2016joint} that concentrate on refining the matching strategy.
The method in~\cite{das2014consistent} focuses on the consistent matching between persons under the camera network, and the method in~\cite{rezatofighi2016joint} optimizes the one-to-one matchings on the image level. 
Nevertheless, they do not fundamentally eliminate the ambiguity, since the matching ambiguity, in essence, originates from mismatchings between the probe persons and the gallery persons on the identity level. 

\begin{table}[tb!]
\begin{center} 
\caption{Statistics of the relationship between the Rank-1 (\%) index of the person matching results and the percentage of matching results with ambiguity $P_{am}$ on VIPeR and DukeMTMC datasets.}
\renewcommand\arraystretch{1.1}
\resizebox{0.4\textwidth}{!}{
\begin{tabular}{l|l||c|c}
\Xhline{1.2pt}
 \multicolumn{1}{c|}{Dataset} & \multicolumn{1}{c||}{Method} & Rank-1 & $P_{am}$ \\
\hline
\multirow{3}*{VIPeR}
 & LOMO+XQDA~\cite{liao2015person} & 40.0 & 60.4 \\
 & GOG~\cite{matsukawa2016hierarchical} & 49.7 & 51.6 \\
 & CRAFT~\cite{chen2018person} & 50.3 & 50.4 \\
 \hline
\multirow{3}*{DukeMTMC}
 & CamStyle~\cite{zhong2018camera} & 78.3 & 72.8 \\
 & BoT~\cite{Luo_2019_CVPR_Workshops} & 86.4 & 36.1 \\
 & PBC~\cite{cao2020} & 91.2 & 33.3 \\
\Xhline{1.2pt}
\end{tabular}
}
\label{tab1}
\end{center}
\end{table}

Imagine the case of the perfect probe-gallery matchings, \emph{i.e.} for all probe images, the corresponding gallery images with the same identity are correctly identified, and there must be no ambiguity in the matching results. The less ambiguous matchings are usually associated with better ReID accuracy, which can be supported by the results in Table~\ref{tab1}. In a way, we might conclude that the essence of performance enhancement is to resolve the ambiguous matching on the person identity level in person ReID.
For this, an intuitive assumption can be set, \emph{i.e.} images belonging to \textbf{the same person identity} should \textbf{not} match with images belonging to \textbf{multiple different person identities} across views, called the \emph{unicity} of the person matching on the identity level. 
Based on this, we aim at achieving the unicity matching on the identity level for improving the performance of ReID in this paper.
Correspondingly, we propose a novel end-to-end context-aided unicity matching framework, in which the contextual information is fully explored in both the feature learning phase and the similarity matching phase to achieve the final unicity matching on the identity level in person ReID.
In the feature learning phase, we compute the sample's feature representation with the consideration of its context samples' information by utilizing the graph neural network, by which an initial soft matching can be computed.
In the similarity matching phase, we further refine the soft matching results by utilizing the samples’ context relationship, thus reaching the unicity of person matching.


Specifically, we translate the person ReID into a bipartite graph matching problem in the similarity matching phase. 
Based on a basic setting in which there is one collected image for each person in the dataset, called one-shot ReID, we adopt the Jonker-Volgenant algorithm~\cite{jonker1987shortest} to solve the bipartite graph matching problem for the unicity matching.
Furthermore, with the full consideration of real-world person ReID applications, in which the same person is usually captured multiple times from the camera views and there are usually multiple images of the same person in the dataset, called multi-shot ReID, we need to solve one critical problem when achieving the unicity matching on the identity level, \emph{i.e.} how to identify the images belonging to the same person in the dataset. 
For this, we decompose the problem into two parts: identifying the same person's images under the same view and identifying that from different views.
For the first one, it is a clustering problem in the homogeneous data space, and we devise a clustering-based merging strategy to aggregate the images of the same person.
For the second one, it is back to a person ReID problem itself, and there is no straightforward solution to this chicken and egg problem. In order to tackle the difficulty, we develop a divide-and-conquer scheme to pursue the unicity matching. 
With the aid of the bipartite graph matching theory, we conduct a globally optimal matching association on the person identity level to keep the unicity of the person matching. 
In a way, we compute the similarity matching relationship between samples from a global context perspective.
Beyond that, we develop a fast version of the unicity matching without the performance penalty to scale well up to large-scale real-world ReID applications.

To summarize, the major contributions of this paper are, 
\begin{enumerate}[(1)]
\item we propose and pursue the unicity matching on the person identity level in Re-ID problem, which relieves the ambiguous matching and boosts the performance;
\item we derive an end-to-end ReID matching architecture by utilizing the contextual information not only in the feature learning phase but also in the similarity matching phase;
\item in view of the real-world applications, we achieve the unicity matching in both one-shot ReID and multi-shot ReID, and further develop its fast version;
\item extensive experiments show the effectiveness of the proposed method, with state-of-the-art performances on all five public ReID benchmarks.
\end{enumerate}

\section{Related Work}
In recent years, there has been a growing research interest on person ReID with varied settings, including image-based, video-based and unsupervised ReID.
We witness a rapid advance of these research works on performance~\cite{he2021transreid,yang2021two,tang2020cgan}.
However, due to the common `selfish' matching strategy, these works tend to obtain the ambiguous matching results.
We propose an end-to-end context-aided person ReID framework, which mainly focuses on exploring matching unicity with full utilization of contextual information. 
In the following, we briefly review the unicity-related person ReID methods and the context-based person ReID methods.

\subsection{Exploring Unicity in Person ReID}
The relational consistency at the semantic level as a strong constraint provides extra information for the matching task in person ReID with recent studies~\cite{shen2015person,zhou2018graph,lin2017consistent-aware,ye2017dynamic,das2014consistent,rezatofighi2016joint}. 
Although they share a similar concept of consistency with the proposed $unicity$, there still exists the essential difference at the specific definition and application level.

The methods in ~\cite{shen2015person,zhou2018graph} establish the one-to-one matching relation for solving cross-view misalignments between images.
In~\cite{lin2017consistent-aware}, the image's feature is learned by the matching consistency-aware information under the camera network in a deep learning framework.
In~\cite{ye2017dynamic}, a bipartite graph is built to link the same person across views for estimating the sample labels and learning a discriminative metric in an unsupervised way.
In these methods, the relational consistency is explored for feature learning or label estimation, and the resulting matching results are still ambiguous due to the neglect of matching unicity.

As the most closely related works with ours, Das \emph{et al.}~\cite{das2014consistent} propose eliminating the ambiguous matching under the camera network by solving a sophisticated integer programming problem.
Rezatofighi \emph{et al.}~\cite{rezatofighi2016joint} aim at approximating the globally optimal one-to-one image matching solution by using the binary tree partitioning based m-best solutions.
By comparison, the proposed method achieves the unicity matching under a problem set of the matchings between two disjoint person sets on the person identity level, which abstracts the real-world scenario better compared to the joint matchings on the entire camera network in~\cite{das2014consistent} and has a more natural and beneficial constraint than the method with an image-based matching consistency in~\cite{rezatofighi2016joint}. 
Besides, the two methods are difficult to scale up to the large-scale ReID due to the high time complexity. We redefine the matching problem under the camera network into several independent sub-matching problems between camera pairs, which is parallel-computing friendly and can scale well up to large-scale ReID applications.

\subsection{Exploring Context in Person ReID}
The contextual information, as the useful complementary information to the individual features, has been well exploited in person ReID~\cite{cao2017from,yan2019learning,chen2018group,liu2021learning,shen2018person}. 
The person image's feature representation is learned  by combining its individual information and its contextual information.
For instance, 
{Shen \emph{et al.}~\cite{shen2018person} estimate the probe-gallery relation feature with considering the relationship information of other pairs.} 
The method in~\cite{luo2019spectral} captures informative connections among images by a proposed spectral feature transformation method; in~\cite{li2019adaptive}, the relationship among the images and their nearest neighbors are fully explored to extract a robust feature representation; 
Ji \emph{et al.}~\cite{ji2020context} propose a context-aware graph convolution network with full exploitation of hard matched images for ReID;
Liu \emph{et al.}~\cite{liu2021learning} construct the locally-connected graph and the fully-connected graph to explore the relationships among person images for ReID. 
Also, re-ranking techniques achieve performance improvement by utilizing contextual information~\cite{sarfraz2018a,wang2019incremental}. 
Zhong \emph{et al.}~\cite{zhong2017re-ranking} advocate that the true matching pairs share similar k-reciprocal nearest neighbors and compute the k-reciprocal features to re-rank the ReID results;
the method in~\cite{cao2020} refines the pairwise similarity along the geodesic path of the underlying data manifold by the contextual information.
By contrast, we entirely use the contextual information in the pairwise similarity computation and person matching phases to achieve performance enhancement.

\begin{figure*}[tb!]
\begin{center}
   \includegraphics[width=1.0\linewidth, height=0.36\linewidth]{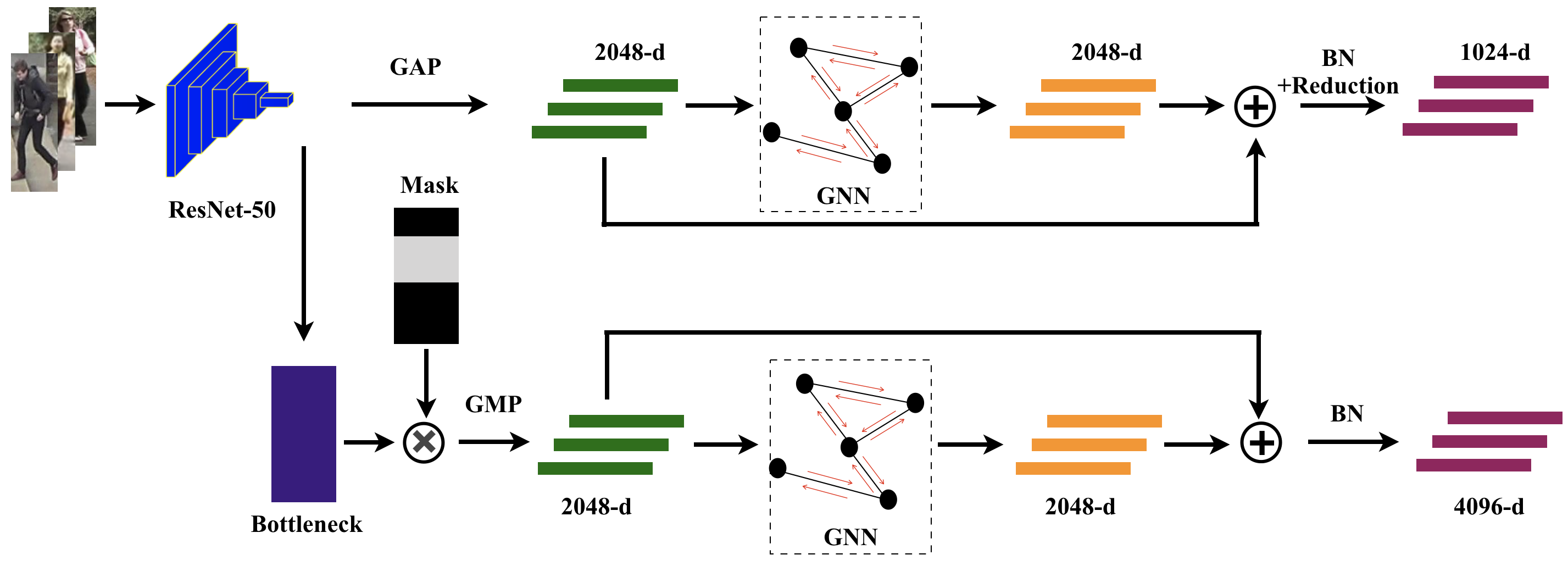}
\end{center}
   \caption{The network structure of soft matching procedure. The GAP, GMP and GNN denote global average pooling, global max pooling and graph neural network, respectively. BN+Reduction represents a batchnorm layer followed by a reduction layer. Due to the information loss on the masked feature map, the dimension-reduced feature is not expressive and there is no reduction in the bottom dropping branch.}
\label{fig2}
\end{figure*}

\section{Methodology}
Given a probe set $\mathcal{P}=\left \{ p_i\:| \: i=1,\cdots ,N\right \}$ with $N$ image samples captured from some camera views, and a gallery set $\mathcal{G}=\left \{ g_i\:| \: i=1,\cdots ,M\right \}$ with $M$ image samples from other views.
The person ReID is to match each probe image sample $p_i$ ($i=1,2,\cdots, N$) from these gallery image samples $g_i$ ($i=1,2,\cdots, M$) by the pairwise similarities. 
Its goal is to achieve as many correct and unique matchings as possible.

To relieve the ambiguous matching, we propose an end-to-end person unicity matching method by online computing initial soft matchings based on the graph neural network firstly and offline refining the soft matchings based on the bipartite graph matching. 

\subsection{Soft Matching by Graph Neural Network}
We model the soft person matching procedure by extracting a discriminative representation, which contains information from similar samples. As illustrated in Fig.~\ref{fig2}, three modules are included in the model: individual feature extraction, context feature extraction and batch dropblock. 

\textbf{Individual feature extraction module.}
We use the ResNet-50~\cite{he2016deep} as the backbone for the individual feature extraction. 
Referring to the BoT method~\cite{Luo_2019_CVPR_Workshops}, we revise the network structure by changing the stride of the last spatial down-sampling operation from $2$ to $1$ to get more fine-grained features. In this way, we obtain a larger feature map of size $2048 \times 16 \times 8$ for the input image size $256 \times 128$. Finally, a $2048$-dimensional individual feature vector is obtained by applying global average pooling.

\textbf{Context feature extraction module.} 
Given a batch with $B$ image samples, we construct a graph $\mathcal{G}=(\mathcal{V},\textbf{A},\textbf{X})$ as the input of graph neural network for the context feature extraction, in which $\mathcal{V}=\left \{1,2,\cdots ,B\right \}$ is the set of sample nodes, $\textbf{A} \in \left \{ 0,1\right \}^{ B \times B }$ is the adjacency matrix and $\textbf{X}=\left [ \bm{x_1}, \cdots, \bm{x_B} \right ]^T \in \mathbb{R}^{B \times \cdot}$ is the node feature matrix. 

Specifically, the adjacency matrix $\textbf{A}$ is computed by searching $k$ nearest neighbors of each sample based on the pairwise similarity from the BoT method\footnote{Instead of adopting the dynamic adjacency matrix from the similarity measurement between individual features during the training process, we adopt a fixed and reliable adjacency matrix from the state-of-the-art BoT~\cite{Luo_2019_CVPR_Workshops} for obtaining powerful contextual features.}~\cite{Luo_2019_CVPR_Workshops}, and the node feature matrix $\textbf{X}$ is composed of the samples' individual features from the individual feature extraction module. 
We perform the graph convolutional operation on the input node features $\textbf{X}$ to generate the nodes' context features~\cite{fey2020deep}:
\begin{equation}\label{e1}
\bm{x_i^{t+1}}=\sigma (\bm{W_1^{t+1}}\bm{x_i^{t}}+\sum_{j\rightarrow i}\bm{W_2^{t+1}}\bm{x_j^{t}}+\sum_{i\rightarrow j}\bm{W_3^{t+1}}\bm{x_j^{t}} ),
\end{equation}
where $\sigma$ denotes ReLU followed by a dropout with probability of $0.5$. The final $2048$-dimensional context features are obtained via $\bm{x'_{i}}=\bm{W_4}[\bm{x_i^1},\cdots,\bm{x_i^T}]$ and $\bm{x_i^1}=\bm{x_i}$ is the input individual feature. 
In view of that too many graph convolutional operations may lead to trivial feature representation for every node and cause negative effect on performance, we set $T=3$ in experiments.

\textbf{Batch dropblock module.}
To further improve soft matching, we include a batch dropblock in which the same region of all feature maps in a batch is dropped randomly to reinforce the attentive feature learning of local regions~\cite{dai2018batch}. 

In summary, the soft person matching is modeled based on a two-branch network composed of a global branch and a dropping branch. 
In each branch, the $2048$-dimensional individual feature and $2048$-dimensional context feature are concatenated as the $4096$-dimensional mixed feature.
In the global branch, the mixed feature is operated by a batchnorm layer and a reduction layer to obtain a $1024$-dimensional feature.
In the dropping branch, a batchnorm layer operates on the mixed feature to obtain a $4096$-dimensional feature. Due to the information loss on the dropped feature maps, there is no reduction in the dropping branch.
The features from the two branches are concatenated as the final feature representation. 
The cosine distance is adopted as the pairwise similarity measure, and then the soft matching results are obtained by searching the maximum of similarity values between each of the probe samples and the gallery samples.


In short, we compute the sample's individual feature vector and further context feature vector in the global branch and the dropping branch, respectively, and combine these features to represent the sample. Based on that, we obtain relatively reliable initial matching results, which approach closer to the unicity matching results than the results from only using the individual feature information.

\subsection{Unicity Matching by Bipartite Graph Matching}
Due to the conventional `selfish' pairwise matching strategy that neglects the global contextual information, there is still ambiguous matching in the soft matching results. We refine the results by relieving the ambiguity and approximating the unicity matching. To achieve this, we transform the person matching problem into a one-to-one bipartite graph matching problem with the probe set and gallery set being two disjoint sets. 

\begin{figure}
\begin{center}
   \includegraphics[width=0.63\linewidth, height=0.7\linewidth]{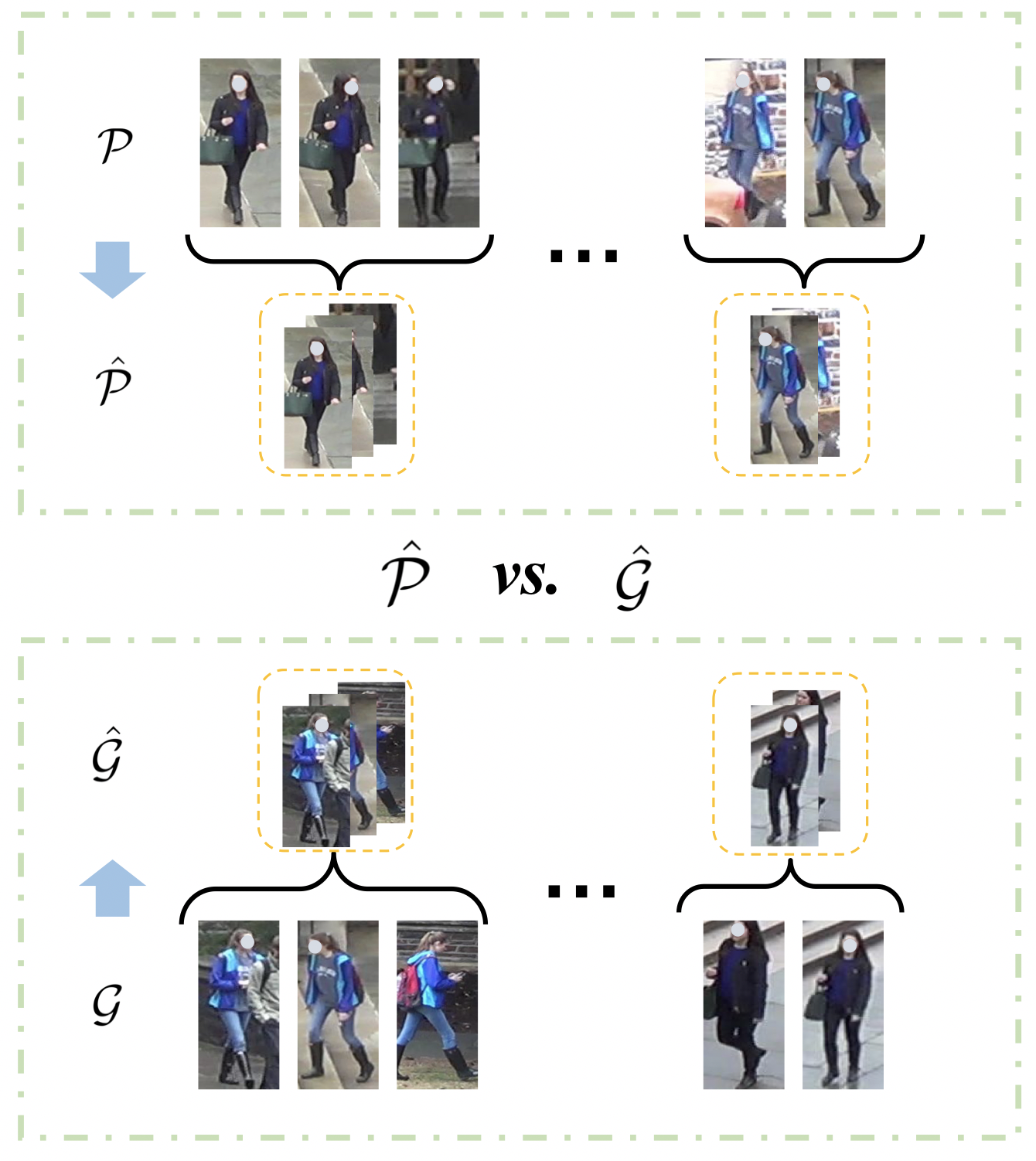}
\end{center}
   \caption{The transformation from the image sets $\mathcal{P}$, $\mathcal{G}$ to the person identity sets $\hat{\mathcal{P}}$, $\hat{\mathcal{G}}$ for the identity-level unicity matching.}
\label{fig5}
\end{figure}

The probe set $\mathcal{P}=\left \{ p_i\:| \: i=1,\cdots ,N\right \}$ and the gallery set $\mathcal{G}=\left \{ g_j\:| \: j=1,\cdots ,M\right \}$ are first rewritten as the probe identity set $\hat{\mathcal{P}}=\left \{ P_u\:| \:u=1,\cdots ,\hat{N}\right \}$ and the gallery identity set $\hat{\mathcal{G}}=\left \{ G_v\:| \: v=1,\cdots ,\hat{M}\right \}$, 
where $P_u$ is the image set of the $u$-th probe identity with ${N}_u$ probe image samples, $G_v$ is the image set of the $v$-th gallery identity with ${M}_v$ gallery image samples.
We illustrate the transformation from the image set to the person identity set in Fig.~\ref{fig5}. 

Assuming that images sharing the same person identity should not match with images with multiple different person identities across views, which is essentially the identity-level unicity matching. We solve the bipartite graph matching problem on the identity-level probe/gallery sets:
\begin{equation}
\begin{aligned}\label{e2}
& \arg\max\limits_{E}\mathop{\sum}\limits_{u=1}^{\hat{N}}\mathop{\sum}\limits_{v=1}^{\hat{M}}e_{uv}s_{uv} \\
& {\rm{s.t}} \quad \forall{u,v} \quad \sum_{u}e_{uv}\leq 1
 \quad \sum_{v}e_{uv}= 1,
\end{aligned}
\end{equation}
where $s_{uv}$ is the similarity between $P_u \in \hat{\mathcal{P}}$ and $G_v \in \hat{\mathcal{G}}$, $e_{uv}\in \left \{ 0,1\right \}$ and $e_{uv}=1$ refers that the $u$-th probe person matches with the $v$-th gallery person, 
$E \in \mathbb{R}^{\hat{N} \times \hat{M}}$ is a sparse matrix composed of $e_{uv}$ ($u=1,\cdots,\hat{N}$, $v=1,\cdots,\hat{M}$).
$\sum_{v}e_{uv}= 1$ represents that there is always an exact and exclusive matched gallery person for each probe person, in turn, $\sum_{u}e_{uv}\leq 1$ means that not every gallery person needs to match with the probe persons\footnote{This paper follows the main stream line of research~\cite{ye2021deep,zhuang2020rethinking,luo2019spectral} addressing the person ReID under the closed-world setting, so each probe person can always find the exact match in the gallery set.}.
The constraints meet the unicity matching.
Our model aims to pick the matching matrix $E$ with maximum total similarity under the constraints, leading to the optimal matchings characterized by unicity.

We address the Eq.\ (\ref{e2}) by the Jonker-Volgenant algorithm~\cite{jonker1987shortest}, which is a variant of the famous Hungarian algorithm~\cite{kuhn1955hungarian} and is much faster for the linear assignment problem in Eq.\ (\ref{e2}).
Note, however, that the input similarity $s_{uv}$ in Eq.\ (\ref{e2}) is identity-based and the resulting matching matrix $E$ is also identity-based matching results, yet the similarity obtained by the soft person matching procedure is image-based and the final output in ReID is also expected to be image-based matching results.
There is a gap between the addressed optimization problem in Eq.\ (\ref{e2}) and the targeting person ReID problem.
The following discussion bridges this gap by the detailed analysis of both dataset configurations, including one-shot and multi-shot settings. 

\subsubsection{Unicity Matching for One-Shot ReID} 
There is only one captured person image for each person in the probe set and gallery set in the one-shot setting, thus $\hat{\mathcal{P}}=\mathcal{P}$ and $\hat{\mathcal{G}}=\mathcal{G}$. 
The identity-based similarity $s_{uv}$ in Eq.\ (\ref{e2}) equals to the image-based similarity obtained by the soft person matching, and the resulting matching matrix $E$ by Eq.\ (\ref{e2}) is actually also the image-based matching result characterized by unicity as the targeting ReID solution.

\subsubsection{Unicity Matching for Multi-Shot ReID} 
There are multiple captured images for each person in the multi-shot setting.
The probe identity set $\hat{\mathcal{P}}$ and the gallery identity set $\hat{\mathcal{G}}$ are firstly obtained by merging the image samples with the same identity. Then we compute the identity-based similarity $s_{uv}$ between the $u$-th probe person in $\hat{\mathcal{P}}$ and the $v$-th gallery person in $\hat{\mathcal{G}}$, and further solve the optimization problem in Eq.\ (\ref{e2}) to match the identity between $\hat{\mathcal{P}}$ and $\hat{\mathcal{G}}$. Based on the obtained identity-based matching results, we compute the final image-based matching results.

In the following, we describe the mergence process, the identity-based matching process and the image-based matching process in detail, respectively.

\textbf{Mergence process.}
Two cases for merging the samples are the image samples with the same identity under the same camera view (SISC sample) and the image samples with the same identity across different camera views (SIDC sample). 
It is worth noting that the merge of SIDC samples can be technically decomposed into two steps: first merging the samples with the same identity within one view into the image set (\emph{i.e.}, the merge of SISC samples) and then merging the image sets with the same identity across views.
We devise \emph{a clustering-based merging strategy} for the merge of SISC samples.
On this basis, we develop \emph{a divide-and-conquer matching strategy} for the merge of samples across views, thus enabling the merge of SIDC samples.

\begin{figure}[t!]
\begin{center}
   \includegraphics[width=1\linewidth, height=0.63\linewidth]{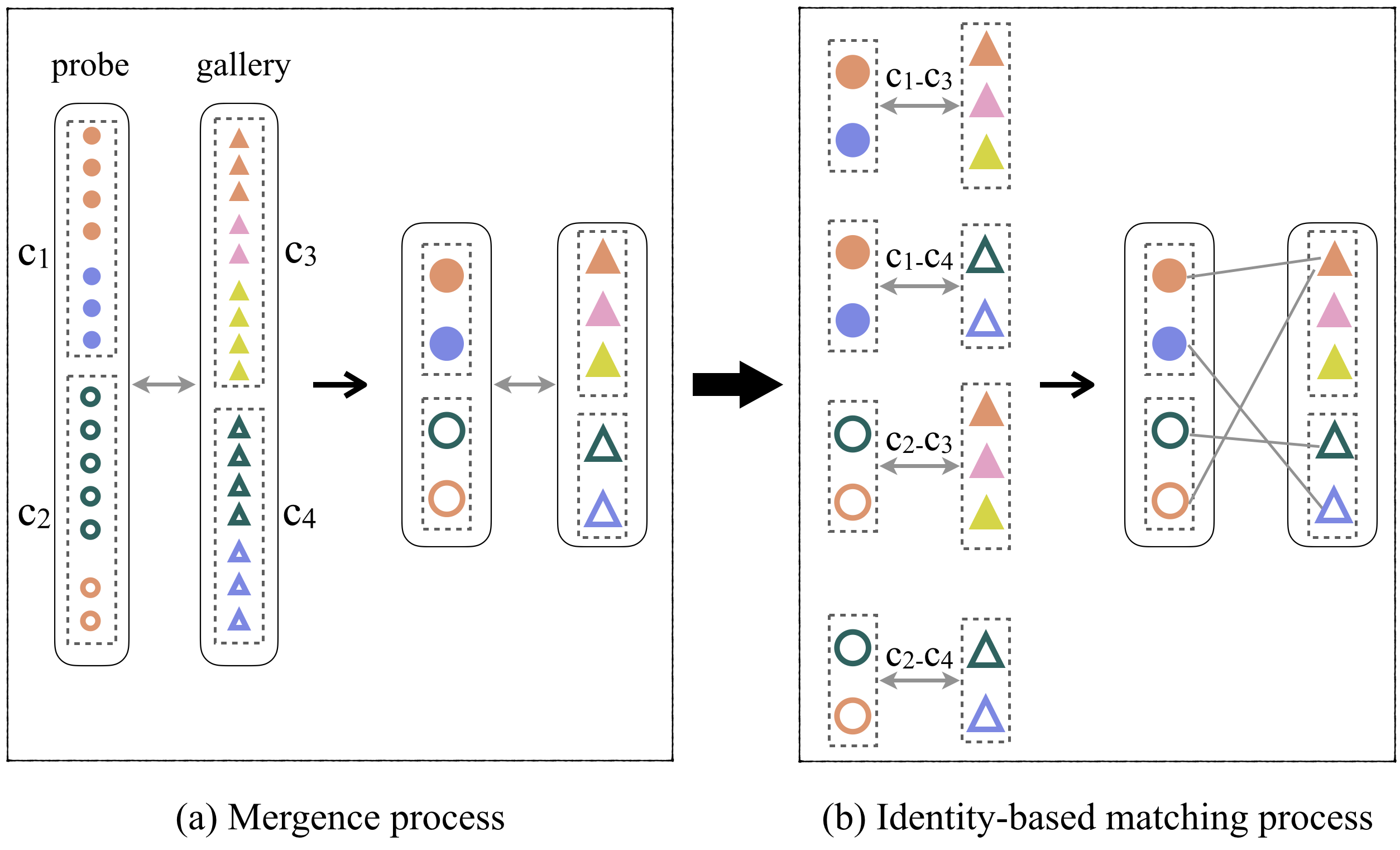}
\end{center}
   \caption{Illustration of the proposed mergence and identity-based matching. The same color represents the same person identity.}
\label{fig3}
\end{figure}

For the merge of SISC samples, it is relatively easy to solve with the help of the spatiotemporal information of the person in reality. When lack of the sample's spatiotemporal information, such as on VIPeR, MSMT17, DukeMTMC, Market1501 and CUHK03 ReID datasets, we adopt the clustering algorithm to achieve the mergence. 
Specifically, taking the data distribution, the algorithm's robustness against parameter variation and its time complexity into consideration, we employ the hierarchical clustering with the agglomerative strategy~\cite{rokach2005clustering}. It is a bottom-up clustering algorithm, in which each sample starts in its own cluster and pairs of clusters are constantly merged as one based on a predefined distance metric, resulting in the final clusters. 

For the merge of SIDC samples, after merging the samples within a camera view, the following mergence across views is in essence a person ReID problem and can not be solved straightforwardly.
However, if we ignore the mergence, the unicity matching will hardly be achieved in the final matching results. 
For this, we propose a divide-and-conquer matching strategy in the following identity-based matching process. 

So far, we have obtained the probe identity sets $\hat{\mathcal{P}}^c$ ($c=1,\cdots ,C_p$) from different cameras and the gallery identity sets $\hat{\mathcal{G}}^c$ ($c=1,\cdots ,C_g$) from different cameras, where $C_p$ and $C_g$ denote the number of cameras capturing the sample images in the probe set and the gallery set, respectively. We illustrate the mergence process in Fig.~\ref{fig3} (a).

\textbf{Identity-based matching process.}
We compute the identity-based similarity $s_{ij}$ between the $i$-th probe person $P_i^c = \left \{ p_u\:| \: u=1,\cdots ,\hat{N}_i^c\right \} \in \hat{\mathcal{P}}^c$ ($c=1,\cdots ,C_p$) and the $j$-th gallery person $G_j^c= \left \{ g_v\:| \: v=1,\cdots ,\hat{M}_j^c\right \} \in \hat{\mathcal{G}}^c$ ($c=1,\cdots ,C_g$) based on the image-based similarity between $p_u$ and $g_v$ computed by the soft matching procedure. It can be viewed as the transformation from the point-to-point measure into the set-to-set measure. The identity-based similarity $s_{ij}$ is computed by picking the largest image-based similarity between $p_u$ ($u=1,\cdots ,\hat{N}_i^c$) and $g_v$ ($v=1,\cdots ,\hat{M}_j^c$). 

The two disjoint sets in the bipartite graph matching in Eq.\ (\ref{e2}) turn into the probe identity set $\hat{\mathcal{P}}^c$ ($c=1,\cdots ,C_p$) associated with camera IDs and the gallery identity set $\hat{\mathcal{G}}^c$ ($c=1,\cdots ,C_g$) associated with camera IDs.
Note that since the SIDC samples are not merged, the probe identities from two different probe identity sets might be the same person, and the same with the gallery identity sets. 
Therefore, the optimization problem in Eq.\ (\ref{e2}) should not be directly solved based on these two disjoint sets, or the unicity matching will not be achieved. For this reason, we propose a divide-and-conquer matching strategy. 
The camera subset pair $(\hat{\mathcal{P}}^{c_p}, \hat{\mathcal{G}}^{c_g})$ $(c_p=1,\cdots ,C_p, c_g=1,\cdots ,C_g, c_p\neq c_g)$ is used to input the optimization problem in Eq.\ (\ref{e2}) for the one-to-one matching, respectively. It can be viewed as a divide step. As a result, for some probe person $P^{c_p}_{i}$ ($c_p=1,\cdots ,C_p, i=1,\cdots ,\hat{N}^{c_p}$), there might exist multiple matching results from different cameras. We pick the match with the largest similarity value as its identity-based matching. It can be viewed as a conquer step.
The merge of SIDC sample is cleverly avoided through the divide-and-conquer strategy, and the unicity matching is well approximated. 
Fig.~\ref{fig3} (b) illustrates the identity-based matching process with the divide-and-conquer strategy.

\textbf{Image-based matching process.}
Based on the obtained identity-based matching results, we compute the image-based matching results. If the hierarchical clustering algorithm correctly clusters all SISC samples, the identity-based matching results equal the image-based matching results. However, there might inevitably exist some wrong clustering results. To make the final matching results approach the global optimum solution, for each identity-based matching $(P_u, G_{v^*})$, we implement the one-to-one matching in Eq.\ (\ref{e2}) to obtain the final image-based matching results. 

\subsection{Acceleration Model for Unicity Matching}
\label{sec:aclr_model}
Based on a common ReID setting~\cite{zhu2020aware,2020Online,fu2019horizontal}, \emph{i.e.} the closed-world setting in which there is always at least one matched sample in the gallery set for each probe sample, we propose a unicity matching model, which is well applied to both one-shot ReID and multi-shot ReID.
Taking a step further, we develop a fast version of the proposed unicity matching to adapt well to large-scale ReID applications.

The unicity matching involves two algorithms for further acceleration: the hierarchical clustering and the Jonker-Volgenant assignment. 
We give three acceleration techniques with almost no accuracy sacrifice.

(1) For the hierarchical clustering, we add a connectivity constraint in which only the interconnected samples can be merged and clustered. 
Specifically, the connectivity constraints are imposed by a connectivity matrix that defines the connected relations between each input sample and its $k_c$ neighboring samples in the data feature space. 
The connectivity constraint makes the clustering algorithm faster with time complexity $O(nk_c)$ instead of $O(n^2)$~\cite{sibson1973slink}, where $n$ is the number of the input samples from one camera.

(2) For the Jonker-Volgenant assignment, we revise the similarity matrix $S$ as a sparse matrix in Eq.\ (\ref{e2}).
Specifically, except for the similarity values between each probe sample and its $k_a$ neighboring gallery samples, the rest in the matrix are set to zero. 
There are only a few non-zero values in the resulting sparse matrix.
As a result, we transform the fully-connected bipartite graph matching problem into a partially connected one.
Given a priori connection information obtained easily and cheaply by the similarity relationship among samples, the assignment algorithm seeks the matching solution faster in a reduced search space.
The resulting time complexity is $O(n^2k_a)$ lower than the original complexity $O(n^2m)$~\cite{jonker1987shortest}, where $n$ and $m$ denote the number of input probe identities from one camera and that of input gallery identities from another camera, respectively.

(3) Due to the camera network configuration, the clustering and assignment algorithms need to run multiple times depending on the number of cameras. 
Owing to the divide-and-conquer strategy, we redefine the complex and large ReID problem under the camera network into several camera-driven independent subproblems, thus we can easily implement our method in camera-driven parallel computing and significantly reduce its time complexity. 


\begin{table}[tb!]
\begin{center}
\caption{Statistics of person ReID datasets. \#TIDs denotes the number of person identity for training, \#PIDs/GIDs denote the number of probe/gallery identities for testing.}
\renewcommand\arraystretch{1.1}
\resizebox{0.4\textwidth}{!}{
\begin{tabular}{p{1.7cm}||p{0.8cm}<{\centering}|p{0.8cm}<{\centering}|p{0.8cm}<{\centering}|c|p{0.6cm}<{\centering}}
\Xhline{1.2pt}
 Dataset & \#TIDs & \#PIDs & \#GIDs & \#images & \#cams \\
\hline
MSMT17  & 1041   & 3060   & 3060  & 126441 & 15 \\
DukeMTMC  & 702   & 702   & 1110   & 36411  & 8 \\
Market1501  & 751   & 750   & 751   & 32668  & 6\\
CUHK03  & 767   & 700  & 700   & 28192  & 2 \\
VIPeR & 316 & 316 & 316 & 1264 & 2 \\
\Xhline{1.2pt}
\end{tabular}
}
\label{tab2}
\end{center}
\end{table}

\section{Experiments}
We conduct experiments on five benchmark person ReID datasets including multi-shot datasets \textbf{MSMT17}~\cite{2018Person}, \textbf{DukeMTMC}~\cite{ristani2016performance}, \textbf{Market1501}~\cite{zheng2015scalable}, \textbf{CUHK03}~\cite{li2014deepreid}, and a one-shot dataset \textbf{VIPeR}~\cite{gray2007evaluating}.

MSMT17 contains 126,441 images which belong to 4,101 identities.
These images are collected with 15 cameras, covering 4 days with different weather conditions over a month. 
The training set consists of 32,621 images of 1,041 identities, whereas the test set contains 93,820 images of the remaining 3,060 identities. 
Compared with other ReID datasets, it is a larger and more challenging dataset. 

DukeMTMC is collected from campus via 8 cameras and is a subset of Duke dataset for the image-based person ReID task. It includes 36,411 images of 1,404 identities, among which the training set consists of 16,522 images of 702 identities and the test set includes 2,228 probe images and 17,661 gallery images of 702 identities.

Market-1501 consists of 32,668 images of 1,501 identities, which are observed in 6 camera views. According to the database setting, 12,936 images from 751 identities are utilized as the training set, while 3,368 probe images and 19,732 gallery images from the remaining 750 identities are used as the test set.

CUHK03 includes 14,097 images of 1,467 identities captured by 6 cameras, of which 7,365 images of 767 identities are used as the training set and 6,732 images of 700 identities are used as the test set. There are 1,400 probe images as well as 5,332 gallery images in the test set. CUHK03 provides both hand-labeled and DPM-detected bounding boxes~\cite{felzenszwalb2009object}. We present the results of both ``labeled'' and ``detected'' data.

VIPeR contains 1,264 images of 632 persons from two non-overlapping camera views in different scenarios. It is a challenging dataset due to its low resolution and high variability in illumination and viewpoint. 

The statistic details of these datasets are summarized in Table~\ref{tab2}. 

\subsection{Implementation and Protocol}

The implementation of the soft matching is conducted on a TITANV GPU with $12$ GB of memory, and that of the unicity matching is conducted on $24$-core Gold5118 CPU with $2.30$ GHz. 
For the soft matching procedure, during training, there are 200 epochs and the batch size is set as $64$ with each identity containing $4$ sample images. The backbone ResNet-50 is initialized from the ImageNet pre-trained model~\cite{deng2009imagenet}. We use the Adam optimizer~\cite{kingma2014adam} with a warmup strategy applied to the learning rate and adopt the triplet loss, center loss and ID loss for training similar to the BoT~\cite{Luo_2019_CVPR_Workshops}. 
During testing, we set the batch size as $128$. 
{The adjacency matrix $\textbf{A}$ is computed based on the BoT method and the number of the neighbor is set to $k=3$ in the context feature extraction module.}
For the unicity matching procedure, there is no model training and a predefined number of cluster in the hierarchical clustering is determined by the Silhouette Coefficient~\cite{zhou2014automatic}. 
The parameters $k_c$ and $k_a$ related to the acceleration model are set to $60$.

We adopt three evaluation metrics to evaluate the effectiveness of the proposed method: cumulated matching characteristics (CMC), mean Average Precision (mAP) and matching unicity measurement ($P_{um}$). $P_{um}=\frac{N_{um}}{N}$ describes the extent of unicity with $N_{um}$ and $N$ denoting the number of the probe images with unicity matching results and the number of the probe images, respectively.

\subsection{Analysis of the proposed method}

\textbf{Effect of two-stage matching architecture.} 
We first conduct experiments on the one-shot VIPeR. Due to the limited number of samples, the soft matching procedure based on deep learning is replaced by the traditional methods and we then perform the unicity matching procedure. As shown in Table~\ref{tab3}, based on different traditional methods, our unicity matching model consistently improves the performance and obtains the perfect unicity matching results. 
Note that the unicity matching results are not necessarily the globally optimal solution. 
Table~\ref{tab3} shows that $100\%$ Rank-1 accuracy cannot be achieved even if $100\%$ $P_{um}$ is obtained. The perfect matching results rely on both the matching unicity and reliable similarity measurements. 
We then test on the multi-shot datasets MSMT17, DukeMTMC, Market1501 and CUHK03(detected). 
As shown in Table~\ref{tab4}, starting from the model with only individual feature extraction as the baseline (w/ IFE), the performance is continuously improved by adding the soft matching procedure (w/ IFE\&CFE\&BD) followed by the unicity matching procedure (w/ SISC\&SIDC) on both datasets. 
Rank-1 accuracy is improved by $7.5\%$ by applying the soft matching and further obtains $21.8\%$ gain by adding the unicity matching on CUHK03(detected). In addition, the consecutive growth of $P_{um}$ indicates that the person unicity matching is progressively achieved by the proposed two-stage matching architecture. 
Moreover, in the unicity matching procedure, we merge the SISC samples by the hierarchical clustering, yet a more accurate mergence can be achieved with the aid of the samples' spatiotemporal information that often exists in practice and then we can obtain better results. 
For reference, we experiment on an ideal case with adopting the ground-truth of SISC samples' mergence in the proposed method on DukeMTMC and CUHK03(detected). The higher Rank-1 accuracies are obtained with $92.7\%$ on DukeMTMC and $89.1\%$ on CUHK03(detected).

\begin{table}[tb!]
\begin{center}
\caption{Performance  (\%) evaluation of the proposed unicity matching with various baseline methods on the VIPeR dataset.}
\renewcommand\arraystretch{1.1}
\resizebox{0.4\textwidth}{!}{
\begin{tabular}{l||c c c}
\Xhline{1.2pt}
 \multicolumn{1}{c||}{Method} & Rank-1  & Rank-10 & $P_{um}$ \\
\hline
LOMO+XQDA~\cite{liao2015person} & 40.0 & 80.5 & 39.6 \\
+Unicity Matching  & 49.2 & 84.8 & 100.0 \\
\hline
GOG~\cite{matsukawa2016hierarchical} & 49.7 & 88.7 & 48.4 \\
+Unicity Matching  & 60.0 & 90.7 & 100.0 \\
\hline
CRAFT~\cite{chen2018person} & 50.3 & 89.6 & 49.6\\
+Unicity Matching  & 59.3 & 92.9 & 100.0 \\
\hline
HP-net~\cite{liu2017hydraplus} & 56.6 & 87.0 & 56.3\\
+Unicity Matching  & 65.2 & 91.4 & 100.0 \\
\Xhline{1.2pt}
\end{tabular}
}
\label{tab3}
\end{center}
\end{table}

\begin{table*}[tb!]
\begin{center}
\caption{Performance (\%) comparison of the proposed method with various settings on MSMT17, DukeMTMC, Market1501 and CUHK03(detected).}
\renewcommand\arraystretch{1.1}
\resizebox{0.9\textwidth}{!}{
\begin{tabular}{c c c | c c ||c c c|c c c|c c c|c c c}
\Xhline{1.2pt}
\multicolumn{5}{c||}{Model} & \multicolumn{3}{c|}{MSMT17} & \multicolumn{3}{c|}{DukeMTMC} & \multicolumn{3}{c|}{Market1501} & \multicolumn{3}{c}{CUHK03(detected)} \\
 \hline
\multicolumn{3}{c|}{Soft Matching} & \multicolumn{2}{c||}{Unicity Matching} & \multirow{2}{*}{Rank-1} & \multirow{2}{*}{mAP} & \multirow{2}{*}{$P_{um}$} & \multirow{2}{*}{Rank-1} & \multirow{2}{*}{mAP} & \multirow{2}{*}{$P_{um}$} & \multirow{2}{*}{Rank-1} & \multirow{2}{*}{mAP} & \multirow{2}{*}{$P_{um}$} & \multirow{2}{*}{Rank-1} & \multirow{2}{*}{mAP} & \multirow{2}{*}{$P_{um}$} \\
 IFE & CFE & BD & SISC & SIDC & & & & & & & & & & & & \\
 \hline
\checkmark & $\times$ & $\times$ & $\times$ & $\times$ & 72.5 & 48.3 & 36.9 & 86.4 & 76.4 & 63.9 & 94.5 & 85.9 & 83.2 & 58.3 & 58.0 & 44.9 \\
 \checkmark & \checkmark & $\times$ & $\times$ & $\times$ & 80.0 & 61.7 & 51.4 & 86.8 & 76.7 & 64.6 & 91.2 & 78.2 & 72.7 & 64.9 & 63.3 & 49.0 \\
  \checkmark & $\times$ & \checkmark & $\times$ & $\times$ & 75.2 & 53.0 & 41.4 & 86.9 & 75.4 & 63.7 & 94.5 & 86.6 & 81.2 & 65.4 & 62.6 & 48.5 \\
 \checkmark & \checkmark & \checkmark & $\times$ & $\times$ & 77.1 & 57.0 & 45.4 & 88.4 & 77.0 & 66.8 & 94.6 & 86.9 & 83.8 & 65.8 & 63.7 & 50.3 \\
 \hline
\checkmark & \checkmark & \checkmark & \checkmark & $\times$ & 82.4 & 66.5 & 53.6 & 83.6 & 75.0 & 52.8 & 79.8 & 74.5 & 58.8 & 87.6 & 79.2 & 72.6 \\
 \checkmark & \checkmark & \checkmark & $\times$ & \checkmark & 83.7 & 66.9 & 52.0 & 92.4 & 80.7 & 66.9 & 96.1 & 88.6 & 85.8 & 81.4 & 70.3 & 70.7 \\
 \checkmark & \checkmark & \checkmark & \checkmark & \checkmark & 83.8 & 67.3 & 53.9 & 91.7 & 79.4 & 67.4 & 95.2 & 87.5 & 85.4 & 87.6 & 79.2 & 72.6 \\
\Xhline{1.2pt}
\end{tabular}
}
\label{tab4}
\end{center}
\end{table*}

\begin{table}
\begin{center}
\caption{The variances from $10$ trials of the proposed soft matching with various settings on DukeMTMC and CUHK03(detected).}
\renewcommand\arraystretch{1.1}
\resizebox{0.48\textwidth}{!}{
\begin{tabular}{c c||c c|c c}
\Xhline{1.2pt}
\multicolumn{2}{c||}{Model} & \multicolumn{2}{c|}{DukeMTMC} & \multicolumn{2}{c}{CUHK03(detected)} \\
\hline
\multicolumn{2}{c||}{Soft Matching} & \multirow{2}{*}{Rank-1} & \multirow{2}{*}{mAP} & \multirow{2}{*}{Rank-1} & \multirow{2}{*}{mAP} \\
Individual\_Concat & Individual\_CFE & & & & \\
\hline
\checkmark & \checkmark & 0.0232 & 0.0018 & 0.0671 & 0.0039\\
$\times$ & \checkmark & 0.0517 & 0.0116 & 0.2449 & 0.0418 \\
\checkmark & $\times$ & 0.0432 & 0.0028 & 0.1418 & 0.0054 \\
$\times$ & $\times$ & 1.0556 & 0.0290 & 0.2604 & 0.0573 \\
\Xhline{1.2pt}
\end{tabular}
}
\label{tab6}
\end{center}
\end{table}

\textbf{Analysis on soft matching procedure.} 
We validate the effects of the context feature extraction module (CFE) and the batch dropblock module (BD) on the proposed soft matching in Table~\ref{tab4}.
The BD module has a positive impact on performance (w/ IFE vs.\ w/ IFE\&BD) on both datasets.
Likewise, the CFE module positively affects the performance (w/ IFE vs.\ w/ IFE\&CFE) on both datasets with different scales except on Market1501.
The results on Market1501 show that the introduce of the contextual information could have an adverse effect on performance when the individual feature extraction export the powerful individual feature representation for ReID, \emph{i.e.}, a relatively strong baseline (w/ IFE).
Alternatively, a combination of CFE and BD modules pushes the performance ahead further (w/ IFE vs.\ w/ IFE\&CFE\&BD) on both datasets.

In addition, in the soft matching procedure, both the individual and context features are extracted to represent the sample.
The introduction of the context feature effectively promotes performance improvement, as shown in Table~\ref{tab4} in the paper.
Furthermore, the sample’s context feature is extracted by exploring its neighbor samples in a batch, possibly bringing the randomness of feature representation to the test samples. 
We repeat $10$ trials on DukeMTMC and CUHK03(detected) and report the variances of the results in Table~\ref{tab6}.
It can be seen that the proposed soft matching procedure (w/ Individual\_Concat\&Individual\_CFE) produces minor variances of the results, indicating little influence brought by the batch randomness.
As the final feature representation is composed of the individual feature and the context feature, the introduction of the individual feature effectively alleviates the impact of the randomness on the final representation.
We remove the individual feature in the final concatenation process (w/o Individual\_Concat) or in the CFE module (w/o Individual\_CFE) or in both of them (w/o Individual\_CFE\&Individual\_Concat), and the increasing variances of results from $10$ trials in Table~\ref{tab6} strongly support the conclusion mentioned above.

In conclusion, the context feature combined with the individual feature contributes to the effectiveness and robustness of the method.

\begin{table}[tb!]
\begin{center}
\caption{Comparison with the re-ranking methods based on two baselines on DukeMTMC and CUHK03(detected) datasets.}
\renewcommand\arraystretch{1.1}
\resizebox{0.48\textwidth}{!}{
\begin{tabular}{l||c c c|c c c}
\Xhline{1.2pt}
\multirow{2}{*}{Model} & \multicolumn{3}{c|}{DukeMTMC} & \multicolumn{3}{c}{CUHK03(detected)} \\
  & Rank-1 & mAP & $P_{um}$ & Rank-1 & mAP & $P_{um}$ \\
\hline
BoT(ResNet50)~\cite{Luo_2019_CVPR_Workshops} & 86.4 & 76.4 & 63.9 & 58.3 & 58.0 & 44.9 \\
+k-RE~\cite{zhong2017re-ranking} & 90.3 & 89.1 & 65.5 & 69.6 & 73.2 & 52.2 \\
+ECN~\cite{sarfraz2018a} & 91.2 & 89.4 & 66.9 & 71.3 & 74.5 & 52.8 \\
+PBC~\cite{cao2020} & 90.7 & 85.6 & 65.8 & 72.1 & 74.9 & 53.1 \\
+Unicity Matching & 91.5 & 79.3 & 67.0 & 85.2 & 77.2 & 71.7 \\
\hline
BoT(IBN-Net50-a)~\cite{Luo_2019_CVPR_Workshops} & 89.0 & 78.8 & 67.3 & 62.3 & 60.6 & 46.6 \\
+k-RE~\cite{zhong2017re-ranking} & 91.7 & 90.2 & 69.7 & 73.4 & 76.3 & 55.3 \\
+ECN~\cite{sarfraz2018a} & 92.2 & 90.6 & 70.1 & 73.2 & 76.1 & 54.3 \\
+PBC~\cite{cao2020} & 91.9 & 86.5 & 70.0 & 74.1 & 76.8 & 56.5 \\
+Unicity Matching & 92.6 & 81.4 & 70.8 & 85.6 & 77.4 & 74.6 \\
\Xhline{1.2pt}
\end{tabular}
}
\label{tab7}
\end{center}
\end{table}

\textbf{Analysis on unicity matching procedure.} 
The merges of SISC and SIDC samples are the key parts for realizing the unicity matching on the multi-shot dataset. 
We show the ablation results of the proposed unicity matching with only SISC samples' mergence (w/ SISC) or with only SIDC samples' mergence (w/ SIDC) in Table~\ref{tab4}. 
When we pursue the unicity matching with only considering SISC samples based on the results from the proposed soft matching, the model (w/ SISC) prompts the performance on MSMT17 and CUHK03(detected), while negatively impacts the performance on DukeMTMC and Market1501.
The clustering algorithm in the model (w/ SISC) tends to generate more incorrect merging results on DukeMTMC and Market1501 with more dispersed data distribution, presumably this causes the performance degradation on DukeMTMC and Market1501. 
When we pursue the unicity matching with only considering SIDC samples based on the soft matching, the model (w/ SIDC) brings performance improvement on all datasets.
The proposed unicity matching model (w/ SISC\&SIDC) with complete matching constraint at the semantic level shows absolute advantage compared to the soft matching model, indicating the effectiveness of the unicity matching with both of the two key parts\footnote{There is no merge of SIDC samples since only two cameras are used for person image collection on CUHK03, the model (w/ SISC) is equivalent to the model (w/ SISC\&SIDC).}.
It is worth noting that our unicity matching model (w/ SISC\&SIDC) is slightly inferior to the model (w/ SIDC) on performance on DukeMTMC and Market1501, which might results from the error introduced by the clustering algorithm.


\begin{figure}[tb!]
\begin{center}
   \includegraphics[width=0.9\linewidth, height=0.38\linewidth]{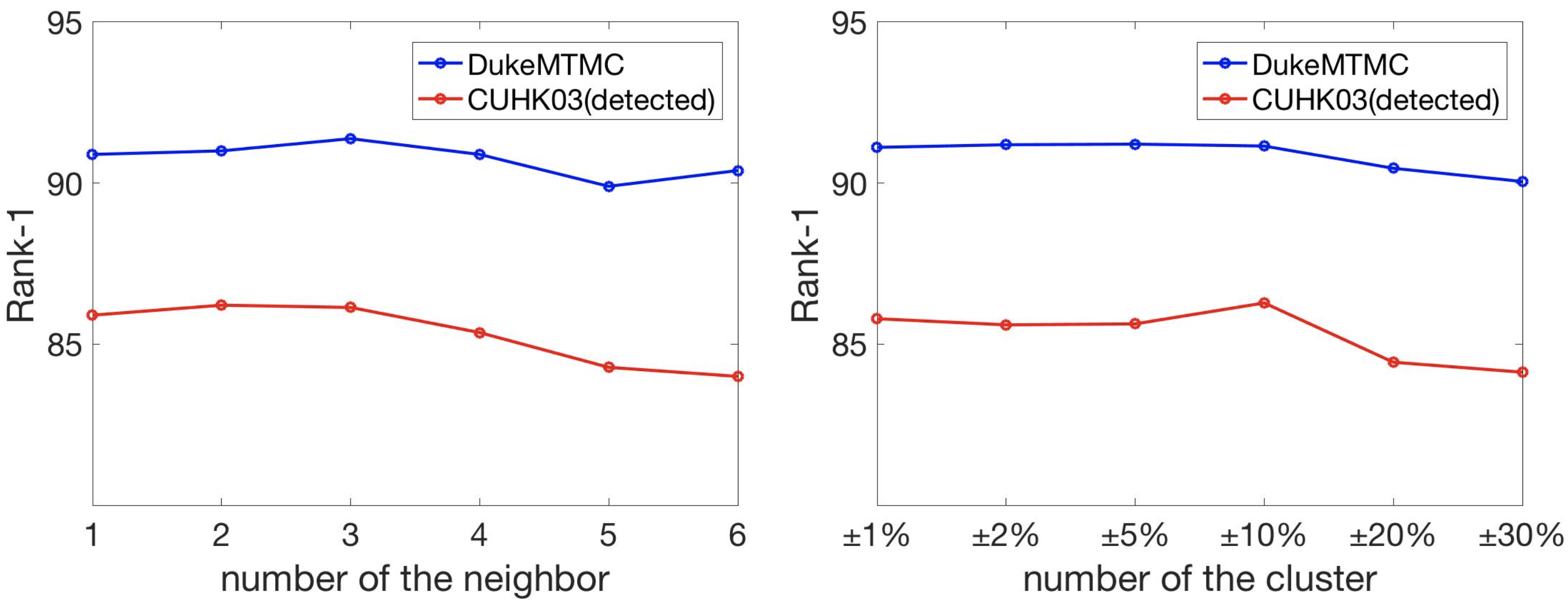}
\end{center}
   \caption{Impact of the number of the neighbor in the soft matching and the number of the cluster in the unicity matching on the performance (\%) on DukeMTMC and CUHK03(detected).}
\label{fig4}
\end{figure}

\begin{figure}[tb!]
\begin{center}
   \includegraphics[width=0.9\linewidth, height=0.73\linewidth]{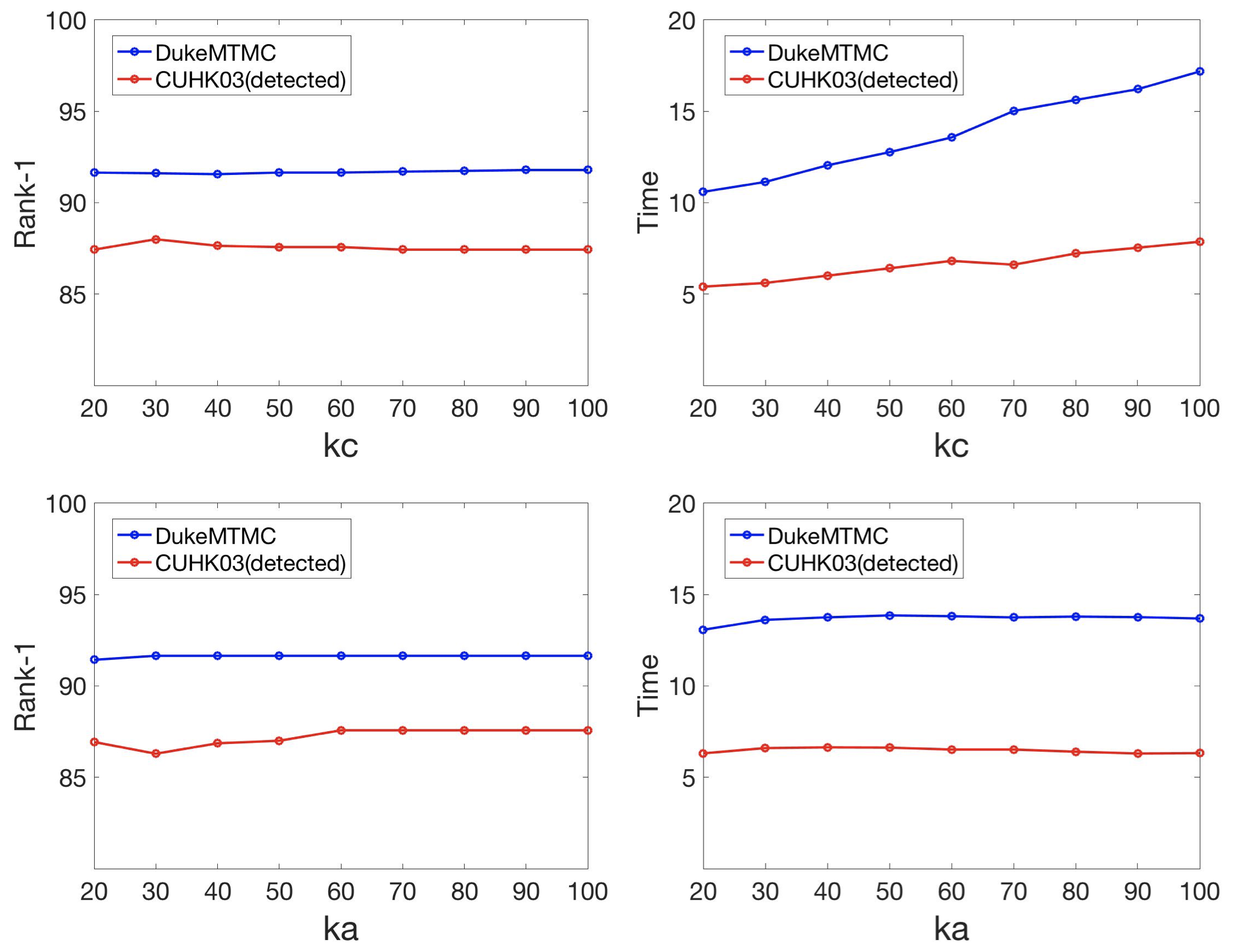}
\end{center}
   \caption{Impact of the parameters $k_c$ and $k_a$ at the Rank-1 performance (\%) and the running time (in seconds) on DukeMTMC and CUHK03(detected).}
\label{fig7}
\end{figure}

\begin{table*}[tb!]
         \begin{center}
         \caption{Comparison results (\%) on MSMT17, DukeMTMC, Market1501 and CUHK03. The best and second results are shown in red/bold and blue/bold, respectively.}
         \renewcommand\arraystretch{1.1}
\resizebox{1\textwidth}{!}{
         \begin{tabular}{l|c||c c | c c | c c | c c | c c}
         \Xhline{1.3pt}
         \multicolumn{2}{c||}{\multirow{2}{*}{Method}} & \multicolumn{2}{c|}{MSMT17} & \multicolumn{2}{c|}{DukeMTMC} & \multicolumn{2}{c|}{Market1501} & \multicolumn{2}{c|}{CUHK03(detected)} & \multicolumn{2}{c}{CUHK03(labeled)}\\
         \multicolumn{2}{c||}{} & Rank-1 & mAP & Rank-1 & mAP & Rank-1 & mAP & Rank-1 & mAP & Rank-1 & mAP \\
         \hline
         BDB+Cut~\cite{dai2018batch} & ICCV2019 & ~~- & ~~- & 89.0 & 76.0 & 95.3 & 86.7 & 76.4 & 73.5 & 79.4 & 76.7 \\
         OSNet~\cite{zhou2019omni} & ICCV2019 & 78.7  & 52.9 & 88.6 & 73.5 & 94.8 & 84.9 & 72.3 & 67.8 & ~~- & ~~- \\
         BAT-net~\cite{fang2019bilinear} & ICCV2019 & 79.5  & 56.8 & 87.7 & 77.3 & 95.1& 87.4 & 76.2 & 73.2 & 78.6 & 76.1 \\
         SFT+LBR~\cite{luo2019spectral} & ICCV2019 & 79.0 & 58.3 & 90.0 & 79.6 & 94.1 & 87.5 & ~~- & ~~- & 74.3 & 71.7\\
         Auto-ReID~\cite{quan2019auto} & ICCV2019 & 78.2 & 52.5 & ~~- & ~~- & 94.5 & 85.1 & 73.3 & 69.3 & 77.9 & 73.0 \\
         ABD-Net~\cite{chen2019abd} & ICCV2019 & 82.3 & 60.8 & 89.0 & 78.6 & 95.6 & 88.3 & ~~- & ~~- & ~~- & ~~- \\
         HPM~\cite{fu2019horizontal} & AAAI2019 & ~~- & ~~- & 86.6 & 74.3 & 94.2 & 82.7 & 63.9 & 57.5 & ~~- & ~~- \\
         ISP~\cite{zhu2020identity} & ECCV2020 & ~~- & ~~- & 89.6 & 80.0 & 95.3 & 88.6 & 75.2 & 71.4 & 76.5 & 74.1 \\
         CBN~\cite{zhuang2020rethinking} & ECCV2020 & 72.8 & 42.9 & 82.5 & 67.3 & 91.3 & 77.3 & ~~- & ~~- & ~~- & ~~- \\
         SAN~\cite{jin2020semantics} & AAAI2020 & 79.2 & 55.7 & 87.9 & 75.5 & 96.1 & 88.0 & 79.4 & 74.6 & 80.1 & 76.4 \\
         VA-reID~\cite{zhu2020aware} & AAAI2020 & ~~- & ~~- & 91.6 & 84.5 & 96.2 & \textcolor[rgb]{0,0,1}{\textbf{91.7}} & ~~- & ~~- & ~~- & ~~- \\
         RGA-SC~\cite{zhang2020relation} & CVPR2020 & 80.3 & 57.5 & ~~- & ~~- & 96.1 & 88.4 & 79.6 & 74.5 & 81.1 & 77.4 \\
         MPN~\cite{ding2020multi} & PAMI2020 & 83.5 & 62.7 & 91.5 & 82.0 & 96.3 & 89.4 & 83.4 & 79.1 & 85.0 & 81.1 \\
         HACNN+DHA-Net~\cite{wang2019learning} & TIP2020 & ~~- & ~~- & 81.3 & 64.1 & 91.3 & 76.0 & ~~- & ~~- & ~~- & ~~- \\
         CDPM~\cite{wang2019cdpm} & TIP2020 & ~~- & ~~- & 90.1 & 80.2 & 95.9 & 87.2 & 78.8 & 73.3 & 81.4 & 77.5 \\
         3D-SF~\cite{chen2021learning} & CVPR2021 & ~~- & ~~- & 88.2 & 76.1 & 95.0 & 87.3 & ~~- & ~~- & ~~- & ~~- \\
         CDNet~\cite{li2021combined} & CVPR2021 & 78.9 & 54.7 & 88.6 & 76.8 & 95.1 & 86.0 & ~~- & ~~- & ~~- & ~~- \\
         AD-SO~\cite{zhang2021coarse} & CVPR2021 & ~~- & ~~- & 87.4 & 74.9 & 94.8 & 87.7 & 81.3 & \textcolor[rgb]{0,0,1}{\textbf{84.1}} & 80.6 & 79.3 \\
         TransReID~\cite{he2021transreid} & ICCV2021 & 86.2 & 69.4 & 90.7 & 82.6 & 95.2 & 89.5 & ~~- & ~~- & ~~- & ~~- \\
         FA-Net~\cite{liu2021end} & TIP2021 & 76.8 & 51.0 & 88.7 & 77.0 & 95.0 & 84.6 & ~~- & ~~- & ~~- & ~~- \\
         \hline
         FFGNN~\cite{li2019adaptive} & ACMMM2019 & ~~- & ~~- & 87.6 & 85.3 & 95.6 & 92.7 & 71.6 & 68.2 & ~~- & ~~- \\
         CAGCN~\cite{ji2020context} & AAAI2021 & ~~- & ~~- & 91.3 & \textcolor[rgb]{0,0,1}{\textbf{85.9}} & 95.9 & 91.7 & ~~- & ~~- & ~~- & ~~- \\
         HRNet~\cite{liu2021learning} & AAAI2021 & 82.0 & 60.6 & 91.9 & 83.7 & \textcolor[rgb]{0,0,1}{\textbf{96.7}} & 91.3 & 81.1 & 78.4 & ~~- & ~~- \\
         HLGAT~\cite{zhang2021person} & CVPR2021 & \textcolor[rgb]{1,0,0}{\textbf{87.2}} & \textcolor[rgb]{1,0,0}{\textbf{73.2}} & \textcolor[rgb]{1,0,0}{\textbf{92.7}}  & \textcolor[rgb]{1,0,0}{\textbf{87.3}} & \textcolor[rgb]{1,0,0}{\textbf{97.5}} & \textcolor[rgb]{1,0,0}{\textbf{93.4}} & 83.5 & 80.6 & ~~- & ~~- \\
         \hline
         ResNet-50~\cite{he2016deep} & Backbone & 72.5 & 48.3 & 86.4 & 76.4 & 94.5 & 85.9 & 58.3 & 58.0 & 63.8 & 61.2 \\
         Ours &  & 83.8 & 67.3 & 91.7 & 79.4 & 95.2 & 87.5 & \textcolor[rgb]{0,0,1}{\textbf{87.6}} & 79.2 & \textcolor[rgb]{0,0,1}{\textbf{89.2}} & 83.0 \\
         \hline
         ResNet-50(LUP)~\cite{fu2021unsupervised} & Backbone & 74.5 & 49.6 & 86.6 & 74.9 & 94.6 & 86.5 & 72.7 & 70.8 & 75.5 & 73.5 \\
         Ours &  & \textcolor[rgb]{0,0,1}{\textbf{86.5}} & \textcolor[rgb]{0,0,1}{\textbf{70.5}} & 91.9 & 80.9 & 95.6 & 88.9 & \textcolor[rgb]{1,0,0}{\textbf{94.5}} & \textcolor[rgb]{1,0,0}{\textbf{89.0}} & \textcolor[rgb]{1,0,0}{\textbf{93.3}} & \textcolor[rgb]{1,0,0}{\textbf{90.3}} \\
         \hline
         IBN-Net50-a~\cite{pan2018two} & Backbone & 77.4 & 54.2 & 90.1 & 79.1 & 95.0 & 88.2 & 62.3 & 60.6 & 65.6 & 63.8 \\
         Ours &  & 84.5 & 68.1 & \textcolor[rgb]{0,0,1}{\textbf{92.5}} & 80.3 & 96.5 & 88.9 & 87.4 & 80.3 & 88.9 & \textcolor[rgb]{0,0,1}{\textbf{83.3}} \\
         \Xhline{1.2pt}
         \end{tabular}
         }     
         \label{tab5}
         \end{center}
         \end{table*}

\begin{table*}
\begin{center}
\caption{Running time (in seconds) of unicity matching on MSMT17, DukeMTMC, Market1501 and CUHK03(detected). ACLR is the abbreviation of acceleration. C, A, P denote the accelerations for the clustering and assignment, and parallel-computing, respectively.}
\renewcommand\arraystretch{1.1}
\resizebox{0.9\textwidth}{!}{
\begin{tabular}{c c c||c c | c c | c c | c c}
\Xhline{1.2pt}
\multicolumn{3}{c||}{Model} & \multicolumn{2}{c|}{MSMT17} & \multicolumn{2}{c|}{DukeMTMC} & \multicolumn{2}{c|}{Market1501} & \multicolumn{2}{c}{CUHK03(detected)} \\
(C)ACLR & (A)ACLR & (P)ACLR & Time(s) & Rank-1 & Time(s) & Rank-1 & Time(s) & Rank-1 & Time(s) & Rank-1 \\
\hline
$\times$ & $\times$ & $\times$ & 2876.452 & 83.7 & 103.397 & 91.8 & 120.712 & 95.2 & 27.661 & 87.4 \\
\checkmark & $\times$ & $\times$ & 772.125 & 83.8 & 47.277 & 91.7 & 69.092 & 95.2 & 10.155 & 87.6 \\
\checkmark & \checkmark & $\times$ & 700.871 & 83.8 & 46.532 & 91.7 & 68.035 & 95.2 & 10.079 & 87.6\\
\checkmark & \checkmark & \checkmark & 203.387 & 83.8 & 13.739 & 91.7 & 16.089 & 95.2 & 6.467 & 87.6 \\
\Xhline{1.2pt}
\end{tabular}
}
\label{tab8}
\end{center}
\end{table*}

In addition, the proposed unicity matching model defines a novel unicity matching constraint for ReID and can also be used as a re-ranking technique to enhance the performance.
We compare our unicity matching model and several classic and widely used re-ranking person ReID methods~\cite{zhong2017re-ranking,sarfraz2018a,cao2020} based on two different baselines in Table~\ref{tab7}. 
Our model achieves superior performance compared to these re-ranking methods at Rank-1 and $P_{um}$ on all two datasets. 
Lower mAP on DukeMTMC is because that our model's target is to relieve the matching ambiguity and reach the matching unicity, which is closely related to the Rank-1 and $P_{um}$ indices than the mAP index in the semantic level and naturally results in the improvement at Rank-1 and $P_{um}$. 
It is worth noting that our model still outperforms the re-ranking methods at all indices by a large margin on CUHK03 (detected). 
In particular, it obtains $13.1\%$ and $11.5\%$ Rank-1 advantages compared to the high-performing re-ranking PBC method~\cite{cao2020} with different baselines.
The baseline performance is relatively poor, leading to more ambiguous matching results on CUHK03. The ambiguity is effectively mitigated by applying our unicity matching model and significant improvement is achieved.



\textbf{Parameters analysis.} 
In our method, there are two parameters: the number of the neighbor during context feature extraction, the number of the cluster during the SISC samples' mergence. Fig.~\ref{fig4} shows the impact of two parameters at Rank-1.
The following observations are drawn from Fig.~\ref{fig4}.
(1) The optimal result is achieved with the number of the neighbor equaling to $3$. 
Although four images of each identity are actually used in each training batch, in neighborhood selection, the slightly smaller number of the neighbor ensures identity purity and avoids outliers. 
(2) Our method with the error range of $10\%$ on the number of the cluster has a little harmful effect on performance, indicating its strong robustness against the number of the cluster.

In the acceleration model, there are two extra parameters $k_c$ and $k_a$. 
They control the search range of solution space for accelerating our unicity matching model and the default values are set to $60$ in the experiments. 
We show the impact of these two parameters at Rank-1 and running time on DukeMTMC and CUHK03(detected) in Fig.~\ref{fig7}. 
Each parameter is changed from $20$ to $100$ based on a default value for another parameter. It can be seen that,
(1) the changes of both $k_c$ and $k_a$ have minimal effect on Rank-1;
(2) the larger the $k_c$, the more running time we need, and with increasing $k_a$, the running time is slightly longer at first and then remains at a certain level. 
Based on consideration of both performance and efficiency, we set these two parameters to $60$ in the experiments.

\subsection{Comparison with state-of-the-art}
The comparison results on four large-scale datasets are shown in Table~\ref{tab5}. 
In addition to the the proposed method with the ResNet-50 backbone pre-trained in ImageNet dataset, we also present both results of the proposed method with the ResNet-50 pre-trained in a large-scale person ReID dataset~\cite{fu2021unsupervised} and that with the IBN-Net50-a backbone pre-trained in ImageNet dataset~\cite{pan2018two}.
Considering that we use the contextual information for improving performance in the proposed method, the context-based methods FFGNN~\cite{li2019adaptive}, CAGCN~\cite{ji2020context}, HRNet~\cite{liu2021learning} and HLGAT~\cite{zhang2021person} are specifically presented in a separate frame.
As we can see, on the one hand, the proposed method consistently performs the best or second results at Rank-1 accuracy with all comparisons on MSMT17, DukeMTMC and CUHK03 datasets.
On the other hand, there is no overbearing advantage of the proposed method on mAP on DukeMTMC and Market1501. 
The proposed method mainly focuses on realizing the matching characterized by unicity, which is closely related to the Rank-1 index in the semantic level, naturally leading to more immediate improvement of Rank-1 than mAP.
HLGAT~\cite{zhang2021person} focuses on learning intra- and inter-local relations for Re-ID and gets the performance advantage on MSMT17, DukeMTMC and CUHK03.
The proposed method works on pursuing the identity-level unicity matching that is commonly neglected in other ReID methods and shows an obvious performance advantage on CUHK03.  


\subsection{Further discussion for real-world scenarios}

\textbf{Running time.}
It is important for a ReID system to have reasonable computation complexity in real-world scenarios. 
We develop three acceleration techniques for our unicity matching model: i) adding the connectivity constraints in the clustering ((C)ACLR), ii) sparsifying the similarity matrix in the assignment ((A)ACLR) and iii) using parallel-computing in the unicity matching ((P)ACLR), thanks to its unique parallel-friendliness nature as discussed in Sec.~\ref{sec:aclr_model}.
The running time of one probe image in Table~\ref{tab8} indicates that our model with these techniques achieves fast matchings with almost lossless accuracies on all four large-scale datasets.

\begin{figure}[tb!]
\begin{center}
   \includegraphics[width=0.9\linewidth, height=0.33\linewidth]{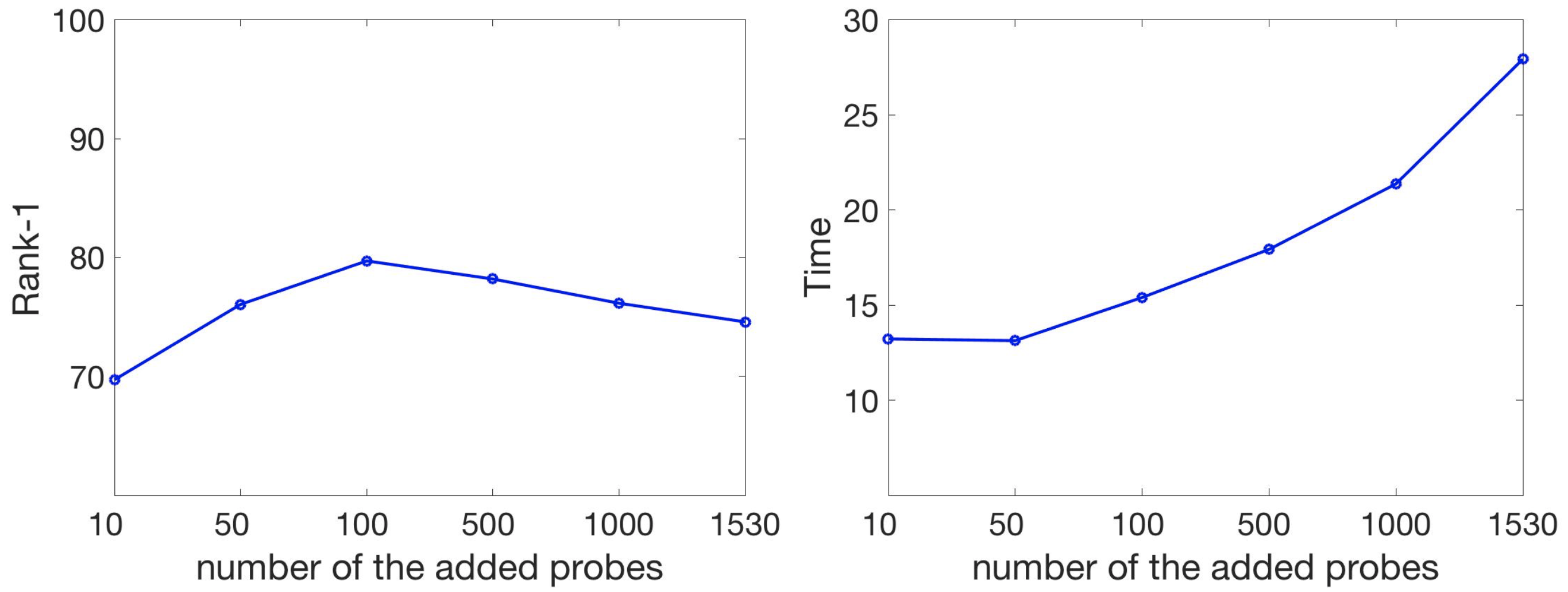}
\end{center}
   \caption{Rank-1 performance (\%) and running time (in seconds) of the proposed unicity matching for the case of adding probes on MSMT17 dataset.}
\label{fig6}
\end{figure}

\textbf{Adding new probes.}
It happens that new probe samples are added in the real-world ReID system. We present a detailed analysis about the feasibility and the complexity of the case.
The proposed unicity matching seeks the optimal matchings with the consideration of all samples in the dataset.
When new probe samples are added in reality, an intuitive solution is to re-run the unicity matching model based on all samples, including the original samples and the newly added samples.
It is, no doubt, a time-consuming solution.
Considering that the added probe samples usually have different identities from the original probe samples in reality and there possibly exist matching errors between the original probe samples and gallery samples, we additionally perform the unicity matching only on the added probe samples and all gallery samples, which is an effective and efficient solution.
We conduct the simulation experiments on the largest dataset MSMT17. 
Half of the probe identities are randomly selected for the original ReID task, and then the samples from the rest of the probes are constantly added into the ReID system as new probe samples.
The Rank-1 accuracy and the running time in Fig.~\ref{fig6} show that the proposed solution is feasible at high efficiency for the real-world case of adding new probes into the ReID system.

\section{Conclusion}
We introduce the matching unicity for person ReID and propose to enhance performance by eliminating the ambiguous matching and pursuing the unicity matching. An end-to-end matching architecture is proposed to learn and refine the person matching results with the contextual information. 
Considering the one-shot and multi-shot datasets, we devise corresponding matching strategies to approximate the matching unicity. 
Specifically, a clustering-based merging strategy and a divide-and-conquer matching strategy are proposed and effectively solve the implementation of the unicity matching on the multi-shot dataset. 
Furthermore, an efficient and fast version of the proposed method with acceleration techniques is developed for scalability.  
Experimental results on five widely used datasets demonstrate the superiority of the proposed method compared to the state-of-the-art ReID methods.

\ifCLASSOPTIONcaptionsoff
  \newpage
\fi



\bibliographystyle{IEEEtran}
\bibliography{egbib}

\begin{thebibliography}{10}
\providecommand{\url}[1]{#1}
\csname url@samestyle\endcsname
\providecommand{\newblock}{\relax}
\providecommand{\bibinfo}[2]{#2}
\providecommand{\BIBentrySTDinterwordspacing}{\spaceskip=0pt\relax}
\providecommand{\BIBentryALTinterwordstretchfactor}{4}
\providecommand{\BIBentryALTinterwordspacing}{\spaceskip=\fontdimen2\font plus
\BIBentryALTinterwordstretchfactor\fontdimen3\font minus
  \fontdimen4\font\relax}
\providecommand{\BIBforeignlanguage}[2]{{%
\expandafter\ifx\csname l@#1\endcsname\relax
\typeout{** WARNING: IEEEtran.bst: No hyphenation pattern has been}%
\typeout{** loaded for the language `#1'. Using the pattern for}%
\typeout{** the default language instead.}%
\else
\language=\csname l@#1\endcsname
\fi
#2}}
\providecommand{\BIBdecl}{\relax}
\BIBdecl

\bibitem{ye2021deep}
M.~Ye, J.~Shen, G.~Lin, T.~Xiang, L.~Shao, and S.~C. Hoi, ``Deep learning for
  person re-identification: A survey and outlook,'' \emph{IEEE Transactions on
  Pattern Analysis and Machine Intelligence}, 2021.

\bibitem{ristani2016performance}
E.~Ristani, F.~Solera, R.~Zou, R.~Cucchiara, and C.~Tomasi, ``Performance
  measures and a data set for multi-target, multi-camera tracking,'' in
  \emph{Proceedings of the European Conference on Computer Vision}.\hskip 1em
  plus 0.5em minus 0.4em\relax Springer, 2016, pp. 17--35.

\bibitem{Luo_2019_CVPR_Workshops}
H.~Luo, Y.~Gu, X.~Liao, S.~Lai, and W.~Jiang, ``Bag of tricks and a strong
  baseline for deep person re-identification,'' in \emph{Proceedings of the
  IEEE Conference on Computer Vision and Pattern Recognition Workshops}, 2019,
  pp. 0--0.

\bibitem{matsukawa2016hierarchical}
T.~Matsukawa, T.~Okabe, E.~Suzuki, and Y.~Sato, ``Hierarchical gaussian
  descriptor for person re-identification,'' in \emph{Proceedings of the IEEE
  Conference on Computer Vision and Pattern Recognition}, 2016, pp. 1363--1372.

\bibitem{RN2020}
H.~Park and B.~Ham, ``Relation network for person re-identification,'' in
  \emph{Proceedings of the AAAI Conference on Artificial Intelligence},
  vol.~34, no.~07, 2020, pp. 11\,839--11\,847.

\bibitem{cao2017from}
M.~Cao, C.~Chen, X.~Hu, and S.~Peng, ``From groups to co-traveler sets: Pair
  matching based person re-identification framework,'' in \emph{Proceedings of
  the IEEE International Conference on Computer Vision Workshops}, 2017, pp.
  2573--2582.

\bibitem{luo2019spectral}
C.~Luo, Y.~Chen, N.~Wang, and Z.~Zhang, ``Spectral feature transformation for
  person re-identification,'' in \emph{Proceedings of the IEEE International
  Conference on Computer Vision}, 2019, pp. 4976--4985.

\bibitem{shen2018person}
Y.~Shen, H.~Li, S.~Yi, D.~Chen, and X.~Wang, ``Person re-identification with
  deep similarity-guided graph neural network,'' in \emph{Proceedings of the
  European conference on computer vision (ECCV)}, 2018, pp. 486--504.

\bibitem{si2018dual}
J.~Si, H.~Zhang, C.-G. Li, J.~Kuen, X.~Kong, A.~C. Kot, and G.~Wang, ``Dual
  attention matching network for context-aware feature sequence based person
  re-identification,'' in \emph{Proceedings of the IEEE Conference on Computer
  Vision and Pattern Recognition}, 2018, pp. 5363--5372.

\bibitem{zhou2018graph}
Q.~Zhou, H.~Fan, S.~Zheng, H.~Su, X.~Li, S.~Wu, and H.~Ling, ``Graph
  correspondence transfer for person re-identification,'' in \emph{Proceedings
  of the AAAI Conference on Artificial Intelligence}, vol.~32, no.~1, 2018.

\bibitem{garcia2017discriminant}
J.~Garcia, N.~Martinel, A.~Gardel, I.~Bravo, G.~L. Foresti, and C.~Micheloni,
  ``Discriminant context information analysis for post-ranking person
  re-identification,'' \emph{IEEE Transactions on Image Processing}, vol.~26,
  no.~4, pp. 1650--1665, 2017.

\bibitem{cao2020}
M.~Cao, C.~Chen, H.~Dou, X.~Hu, S.~Peng, and A.~Kuijper, ``Progressive
  bilateral-context driven model for post-processing person
  re-identification,'' \emph{IEEE Transactions on Multimedia}, vol.~23, pp.
  1239--1251, 2020.

\bibitem{das2014consistent}
A.~Das, A.~Chakraborty, and A.~K. Roy-Chowdhury, ``Consistent re-identification
  in a camera network,'' in \emph{European conference on computer
  vision}.\hskip 1em plus 0.5em minus 0.4em\relax Springer, 2014, pp. 330--345.

\bibitem{rezatofighi2016joint}
S.~Hamid~Rezatofighi, A.~Milan, Z.~Zhang, Q.~Shi, A.~Dick, and I.~Reid, ``Joint
  probabilistic matching using m-best solutions,'' in \emph{Proceedings of the
  IEEE Conference on Computer Vision and Pattern Recognition}, 2016, pp.
  136--145.

\bibitem{liao2015person}
S.~Liao, Y.~Hu, X.~Zhu, and S.~Z. Li, ``Person re-identification by local
  maximal occurrence representation and metric learning,'' in \emph{Proceedings
  of the IEEE Conference on Computer Vision and Pattern Recognition}, 2015, pp.
  2197--2206.

\bibitem{chen2018person}
Y.-C. Chen, X.~Zhu, W.-S. Zheng, and J.-H. Lai, ``Person re-identification by
  camera correlation aware feature augmentation,'' \emph{IEEE Transactions on
  Pattern Analysis and Machine Intelligence}, vol.~40, no.~2, pp. 392--408,
  2017.

\bibitem{zhong2018camera}
Z.~Zhong, L.~Zheng, Z.~Zheng, S.~Li, and Y.~Yang, ``Camera style adaptation for
  person re-identification,'' in \emph{Proceedings of the IEEE Conference on
  Computer Vision and Pattern Recognition}, 2018, pp. 5157--5166.

\bibitem{jonker1987shortest}
R.~Jonker and A.~Volgenant, ``A shortest augmenting path algorithm for dense
  and sparse linear assignment problems,'' \emph{Computing}, vol.~38, no.~4,
  pp. 325--340, 1987.

\bibitem{he2021transreid}
S.~He, H.~Luo, P.~Wang, F.~Wang, H.~Li, and W.~Jiang, ``Transreid:
  Transformer-based object re-identification,'' in \emph{Proceedings of the
  IEEE/CVF International Conference on Computer Vision}, 2021, pp.
  15\,013--15\,022.

\bibitem{yang2021two}
X.~Yang, L.~Liu, N.~Wang, and X.~Gao, ``A two-stream dynamic pyramid
  representation model for video-based person re-identification,'' \emph{IEEE
  Transactions on Image Processing}, vol.~30, pp. 6266--6276, 2021.

\bibitem{tang2020cgan}
Y.~Tang, Y.~Xi, N.~Wang, B.~Song, and X.~Gao, ``Cgan-tm: A novel
  domain-to-domain transferring method for person re-identification,''
  \emph{IEEE Transactions on Image Processing}, vol.~29, pp. 5641--5651, 2020.

\bibitem{shen2015person}
Y.~Shen, W.~Lin, J.~Yan, M.~Xu, J.~Wu, and J.~Wang, ``Person re-identification
  with correspondence structure learning,'' in \emph{Proceedings of the IEEE
  International Conference on Computer Vision}, 2015, pp. 3200--3208.

\bibitem{lin2017consistent-aware}
J.~Lin, L.~Ren, J.~Lu, J.~Feng, and J.~Zhou, ``Consistent-aware deep learning
  for person re-identification in a camera network,'' in \emph{Proceedings of
  the IEEE Conference on Computer Vision and Pattern Recognition}, 2017, pp.
  5771--5780.

\bibitem{ye2017dynamic}
M.~Ye, A.~J. Ma, L.~Zheng, J.~Li, and P.~C. Yuen, ``Dynamic label graph
  matching for unsupervised video re-identification,'' in \emph{Proceedings of
  the IEEE International Conference on Computer Vision}, 2017, pp. 5142--5150.

\bibitem{yan2019learning}
Y.~Yan, Q.~Zhang, B.~Ni, W.~Zhang, M.~Xu, and X.~Yang, ``Learning context graph
  for person search,'' in \emph{Proceedings of the IEEE Conference on Computer
  Vision and Pattern Recognition}, 2019, pp. 2158--2167.

\bibitem{chen2018group}
D.~Chen, D.~Xu, H.~Li, N.~Sebe, and X.~Wang, ``Group consistent similarity
  learning via deep crf for person re-identification,'' in \emph{Proceedings of
  the IEEE Conference on Computer Vision and Pattern Recognition}, 2018, pp.
  8649--8658.

\bibitem{liu2021learning}
S.~Liu, W.~Huang, and Z.~Zhang, ``Learning hybrid relationships for person
  re-identification,'' in \emph{Proceedings of the AAAI Conference on
  Artificial Intelligence}, vol.~35, no.~3, 2021, pp. 2172--2179.

\bibitem{li2019adaptive}
Y.~Li, H.~Yao, L.~Duan, H.~Yao, and C.~Xu, ``Adaptive feature fusion via graph
  neural network for person re-identification,'' in \emph{Proceedings of the
  27th ACM International Conference on Multimedia}, 2019, pp. 2115--2123.

\bibitem{ji2020context}
D.~Ji, H.~Wang, H.~Hu, W.~Gan, W.~Wu, and J.~Yan, ``Context-aware graph
  convolution network for target re-identification,'' \emph{arXiv preprint
  arXiv:2012.04298}, 2020.

\bibitem{sarfraz2018a}
M.~S. Sarfraz, A.~Schumann, A.~Eberle, and R.~Stiefelhagen, ``A pose-sensitive
  embedding for person re-identification with expanded cross neighborhood
  re-ranking,'' in \emph{Proceedings of the IEEE conference on computer vision
  and pattern recognition}, 2018, pp. 420--429.

\bibitem{wang2019incremental}
Z.~Wang, J.~Jiang, Y.~Yu, and S.~Satoh, ``Incremental re-identification by
  cross-direction and cross-ranking adaption,'' \emph{IEEE Transactions on
  Multimedia}, vol.~21, no.~9, pp. 2376--2386, 2019.

\bibitem{zhong2017re-ranking}
Z.~Zhong, L.~Zheng, D.~Cao, and S.~Li, ``Re-ranking person re-identification
  with k-reciprocal encoding,'' in \emph{Proceedings of the IEEE Conference on
  Computer Vision and Pattern Recognition}, 2017, pp. 1318--1327.

\bibitem{he2016deep}
K.~He, X.~Zhang, S.~Ren, and J.~Sun, ``Deep residual learning for image
  recognition,'' in \emph{Proceedings of the IEEE Conference on Computer Vision
  and Pattern Recognition}, 2016, pp. 770--778.

\bibitem{fey2020deep}
M.~Fey, J.~E. Lenssen, C.~Morris, J.~Masci, and N.~M. Kriege, ``Deep graph
  matching consensus,'' \emph{arXiv preprint arXiv:2001.09621}, 2020.

\bibitem{dai2018batch}
Z.~Dai, M.~Chen, X.~Gu, S.~Zhu, and P.~Tan, ``Batch dropblock network for
  person re-identification and beyond,'' in \emph{Proceedings of the IEEE
  International Conference on Computer Vision}, 2019, pp. 3691--3701.

\bibitem{zhuang2020rethinking}
Z.~Zhuang, L.~Wei, L.~Xie, T.~Zhang, H.~Zhang, H.~Wu, H.~Ai, and Q.~Tian,
  ``Rethinking the distribution gap of person re-identification with
  camera-based batch normalization,'' in \emph{Proceedings of the European
  Conference on Computer Vision}, 2020.

\bibitem{kuhn1955hungarian}
H.~W. Kuhn, ``The hungarian method for the assignment problem,'' \emph{Naval
  research logistics quarterly}, vol.~2, no. 1-2, pp. 83--97, 1955.

\bibitem{rokach2005clustering}
L.~Rokach and O.~Maimon, ``Clustering methods,'' in \emph{Data mining and
  knowledge discovery handbook}.\hskip 1em plus 0.5em minus 0.4em\relax
  Springer, 2005, pp. 321--352.

\bibitem{zhu2020aware}
Z.~Zhu, X.~Jiang, F.~Zheng, X.~Guo, F.~Huang, X.~Sun, and W.~Zheng, ``Aware
  loss with angular regularization for person re-identification,'' in
  \emph{Proceedings of the AAAI Conference on Artificial Intelligence},
  vol.~34, no.~07, 2020, pp. 13\,114--13\,121.

\bibitem{2020Online}
J.~Zhou, B.~Su, and Y.~Wu, ``Online joint multi-metric adaptation from frequent
  sharing-subset mining for person re-identification,'' in \emph{Proceedings of
  the IEEE Conference on Computer Vision and Pattern Recognition}, 2020, pp.
  2909--2918.

\bibitem{fu2019horizontal}
Y.~Fu, Y.~Wei, Y.~Zhou, H.~Shi, G.~Huang, X.~Wang, Z.~Yao, and T.~Huang,
  ``Horizontal pyramid matching for person re-identification,'' in
  \emph{Proceedings of the AAAI Conference on Artificial Intelligence},
  vol.~33, 2019, pp. 8295--8302.

\bibitem{sibson1973slink}
R.~Sibson, ``Slink: an optimally efficient algorithm for the single-link
  cluster method,'' \emph{The computer journal}, vol.~16, no.~1, pp. 30--34,
  1973.

\bibitem{2018Person}
L.~Wei, S.~Zhang, W.~Gao, and Q.~Tian, ``Person transfer gan to bridge domain
  gap for person re-identification,'' in \emph{Proceedings of the IEEE
  Conference on Computer Vision and Pattern Recognition}, 2018, pp. 79--88.

\bibitem{zheng2015scalable}
L.~Zheng, L.~Shen, L.~Tian, S.~Wang, J.~Wang, and Q.~Tian, ``Scalable person
  re-identification: A benchmark,'' in \emph{Proceedings of the IEEE
  International Conference on Computer Vision}, 2015, pp. 1116--1124.

\bibitem{li2014deepreid}
W.~Li, R.~Zhao, T.~Xiao, and X.~Wang, ``Deepreid: Deep filter pairing neural
  network for person re-identification,'' in \emph{Proceedings of the IEEE
  Conference on Computer Vision and Pattern Recognition}, 2014, pp. 152--159.

\bibitem{gray2007evaluating}
D.~Gray, S.~Brennan, and H.~Tao, ``Evaluating appearance models for
  recognition, reacquisition, and tracking,'' in \emph{Proc. IEEE international
  workshop on performance evaluation for tracking and surveillance (PETS)},
  vol.~3.\hskip 1em plus 0.5em minus 0.4em\relax Citeseer, 2007, pp. 1--7.

\bibitem{felzenszwalb2009object}
P.~F. Felzenszwalb, R.~B. Girshick, D.~McAllester, and D.~Ramanan, ``Object
  detection with discriminatively trained part-based models,'' \emph{IEEE
  transactions on pattern analysis and machine intelligence}, vol.~32, no.~9,
  pp. 1627--1645, 2009.

\bibitem{deng2009imagenet}
J.~Deng, W.~Dong, R.~Socher, L.-J. Li, K.~Li, and L.~Fei-Fei, ``Imagenet: A
  large-scale hierarchical image database,'' in \emph{IEEE Conference on
  Computer Vision and Pattern Recognition}.\hskip 1em plus 0.5em minus
  0.4em\relax Ieee, 2009, pp. 248--255.

\bibitem{kingma2014adam}
D.~P. Kingma and J.~Ba, ``Adam: A method for stochastic optimization,''
  \emph{arXiv preprint arXiv:1412.6980}, 2014.

\bibitem{zhou2014automatic}
H.~B. Zhou and J.~T. Gao, ``Automatic method for determining cluster number
  based on silhouette coefficient,'' in \emph{Advanced Materials Research},
  vol. 951.\hskip 1em plus 0.5em minus 0.4em\relax Trans Tech Publ, 2014, pp.
  227--230.

\bibitem{liu2017hydraplus}
X.~Liu, H.~Zhao, M.~Tian, L.~Sheng, J.~Shao, S.~Yi, J.~Yan, and X.~Wang,
  ``Hydraplus-net: Attentive deep features for pedestrian analysis,'' in
  \emph{Proceedings of the IEEE international conference on computer vision},
  2017, pp. 350--359.

\bibitem{zhou2019omni}
K.~Zhou, Y.~Yang, A.~Cavallaro, and T.~Xiang, ``Omni-scale feature learning for
  person re-identification,'' in \emph{Proceedings of the IEEE International
  Conference on Computer Vision}, 2019, pp. 3702--3712.

\bibitem{fang2019bilinear}
P.~Fang, J.~Zhou, S.~K. Roy, L.~Petersson, and M.~Harandi, ``Bilinear attention
  networks for person retrieval,'' in \emph{Proceedings of the IEEE
  International Conference on Computer Vision}, 2019, pp. 8030--8039.

\bibitem{quan2019auto}
R.~Quan, X.~Dong, Y.~Wu, L.~Zhu, and Y.~Yang, ``Auto-reid: Searching for a
  part-aware convnet for person re-identification,'' in \emph{Proceedings of
  the IEEE International Conference on Computer Vision}, 2019, pp. 3750--3759.

\bibitem{chen2019abd}
T.~Chen, S.~Ding, J.~Xie, Y.~Yuan, W.~Chen, Y.~Yang, Z.~Ren, and Z.~Wang,
  ``Abd-net: Attentive but diverse person re-identification,'' in
  \emph{Proceedings of the IEEE International Conference on Computer Vision},
  2019, pp. 8351--8361.

\bibitem{zhu2020identity}
K.~Zhu, H.~Guo, Z.~Liu, M.~Tang, and J.~Wang, ``Identity-guided human semantic
  parsing for person re-identification,'' \emph{arXiv preprint
  arXiv:2007.13467}, 2020.

\bibitem{jin2020semantics}
X.~Jin, C.~Lan, W.~Zeng, G.~Wei, and Z.~Chen, ``Semantics-aligned
  representation learning for person re-identification.'' in \emph{Proceedings
  of the AAAI Conference on Artificial Intelligence}, 2020, pp.
  11\,173--11\,180.

\bibitem{zhang2020relation}
Z.~Zhang, C.~Lan, W.~Zeng, X.~Jin, and Z.~Chen, ``Relation-aware global
  attention for person re-identification,'' in \emph{Proceedings of the IEEE
  Conference on Computer Vision and Pattern Recognition}, 2020, pp. 3186--3195.

\bibitem{ding2020multi}
C.~Ding, K.~Wang, P.~Wang, and D.~Tao, ``Multi-task learning with coarse priors
  for robust part-aware person re-identification,'' \emph{IEEE Transactions on
  Pattern Analysis and Machine Intelligence}, 2020.

\bibitem{wang2019learning}
Z.~Wang, J.~Jiang, Y.~Wu, M.~Ye, X.~Bai, and S.~Satoh, ``Learning sparse and
  identity-preserved hidden attributes for person re-identification,''
  \emph{IEEE Transactions on Image Processing}, vol.~29, pp. 2013--2025, 2019.

\bibitem{wang2019cdpm}
K.~Wang, C.~Ding, S.~J. Maybank, and D.~Tao, ``Cdpm: Convolutional deformable
  part models for semantically aligned person re-identification,'' \emph{IEEE
  Transactions on Image Processing}, vol.~29, pp. 3416--3428, 2019.

\bibitem{chen2021learning}
J.~Chen, X.~Jiang, F.~Wang, J.~Zhang, F.~Zheng, X.~Sun, and W.-S. Zheng,
  ``Learning 3d shape feature for texture-insensitive person
  re-identification,'' in \emph{Proceedings of the IEEE/CVF Conference on
  Computer Vision and Pattern Recognition}, 2021, pp. 8146--8155.

\bibitem{li2021combined}
H.~Li, G.~Wu, and W.-S. Zheng, ``Combined depth space based architecture search
  for person re-identification,'' in \emph{Proceedings of the IEEE/CVF
  Conference on Computer Vision and Pattern Recognition}, 2021, pp. 6729--6738.

\bibitem{zhang2021coarse}
A.~Zhang, Y.~Gao, Y.~Niu, W.~Liu, and Y.~Zhou, ``Coarse-to-fine person
  re-identification with auxiliary-domain classification and second-order
  information bottleneck,'' in \emph{Proceedings of the IEEE/CVF Conference on
  Computer Vision and Pattern Recognition}, 2021, pp. 598--607.

\bibitem{liu2021end}
Y.~Liu, W.~Zhou, J.~Liu, G.-J. Qi, Q.~Tian, and H.~Li, ``An end-to-end
  foreground-aware network for person re-identification,'' \emph{IEEE
  Transactions on Image Processing}, vol.~30, pp. 2060--2071, 2021.

\bibitem{zhang2021person}
Z.~Zhang, H.~Zhang, and S.~Liu, ``Person re-identification using heterogeneous
  local graph attention networks,'' in \emph{Proceedings of the IEEE/CVF
  Conference on Computer Vision and Pattern Recognition}, 2021, pp.
  12\,136--12\,145.

\bibitem{fu2021unsupervised}
D.~Fu, D.~Chen, J.~Bao, H.~Yang, L.~Yuan, L.~Zhang, H.~Li, and D.~Chen,
  ``Unsupervised pre-training for person re-identification,'' in
  \emph{Proceedings of the IEEE/CVF Conference on Computer Vision and Pattern
  Recognition}, 2021, pp. 14\,750--14\,759.

\bibitem{pan2018two}
X.~Pan, P.~Luo, J.~Shi, and X.~Tang, ``Two at once: Enhancing learning and
  generalization capacities via ibn-net,'' in \emph{Proceedings of the European
  Conference on Computer Vision}, 2018, pp. 464--479.

\end{thebibliography}
%



%




\end{document}